\documentclass[a4paper]{article}
\usepackage{amsmath}
\usepackage{geometry}
\usepackage{pgf}
\usepackage{tikz}  
\usetikzlibrary{arrows,shapes,chains}  
\usepackage{graphicx}
\usepackage{amssymb}
\usepackage{amsthm}
\usepackage{floatrow}
\usepackage{amsthm}
\usepackage{mathrsfs}
\usepackage{enumitem}
\usepackage{subfigure}
\usepackage{makecell}
\usepackage{appendix}
\usepackage{multirow}
\usepackage{algorithm}
\usepackage{color}
\usepackage{textcomp}
\usepackage{bm}
\usepackage{cite}
\makeatletter
\@addtoreset{equation}{section}
\makeatother

\usepackage{booktabs}
\usepackage{algorithm,algpseudocode,float}
\usepackage{lipsum}
\newtheorem{theorem}{Theorem}[section]
\newtheorem{lemma}{Lemma}[section]

\newtheorem{definition}{\textbf{Definition}}[section]
\newtheorem{remark}{Remark}
\DeclareMathOperator*{\argmin}{argmin}

\makeatletter
\newenvironment{breakablealgorithm}
{% \begin{breakablealgorithm}
	\begin{center}
		\refstepcounter{algorithm}% New algorithm
		\hrule height.8pt depth0pt \kern2pt% \@fs@pre for \@fs@ruled
		\renewcommand{\caption}[2][\relax]{% Make a new \caption
			{\raggedright\textbf{\ALG@name~\thealgorithm} ##2\par}%
			\ifx\relax##1\relax % #1 is \relax
			\addcontentsline{loa}{algorithm}{\protect\numberline{\thealgorithm}##2}%
			\else % #1 is not \relax
			\addcontentsline{loa}{algorithm}{\protect\numberline{\thealgorithm}##1}%
			\fi
			\kern2pt\hrule\kern2pt
		}
	}{% \end{breakablealgorithm}
		\kern2pt\hrule\relax% \@fs@post for \@fs@ruled
	\end{center}
}
\makeatother

\title{\bf T-ADAF: Adaptive Data Augmentation Framework for Image Classification Network based on Tensor T-product Operator}
\author{Feiyang Han \footnote{E-mail: 19110180030@fudan.edu.cn. School of Mathematical Sciences, Fudan University, Shanghai, 200433, P. R.  China. This author is supported by the National Natural Science Foundation of China under grant 12271088. } 
\quad  Yun Miao \footnote{Email: miaoyun2@huawei.com. Theory Lab, Central Research Institute 2012 Labs, Huawei Technologies Co., Ltd. Shanghai, P. R. China.}
\quad  Zhaoyi Sun \footnote{Email: sun.zhaoyi2@huawei.com. Theory Lab, Hong Kong R\&D Center, Huawei Technologies Co., Ltd. Hong Kong, P. R. China.}
\quad  Yimin Wei \footnote{Corresponding author (Yimin Wei). E-mail: ymwei@fudan.edu.cn and yimin.wei@gmail.com. School of Mathematical Sciences and  Shanghai Key Laboratory of Contemporary Applied Mathematics, Fudan University, Shanghai, 200433, P. R. China. This author is supported by Innovation Program of Shanghai Municipal Education Commission and the National Natural Science Foundation of China under grant 12271088.
}
}

\begin{document}

\maketitle

\begin{abstract}
Image classification is one of the most fundamental tasks in Computer Vision. In practical applications, the datasets are usually not as abundant as those in the laboratory and simulation, which is always called as Data Hungry. How to extract the information of data more completely and effectively is very important. Therefore, an Adaptive Data Augmentation Framework based on the tensor T-product Operator is proposed in this paper, to triple one image data to be trained and gain the result from all these three images together with only less than 0.1\% increase in the number of parameters. At the same time, this framework serves the functions of column image embedding and global feature intersection, enabling the model to obtain information in not only spatial but frequency domain, and thus improving the prediction accuracy of the model. Numerical experiments have been designed for several models, and the results demonstrate the effectiveness of this adaptive framework. Numerical experiments show that our data augmentation framework can improve the performance of original neural network model by  2\%, which provides competitive results to state-of-the-art methods.\\



\end{abstract}


% \fi
%\IEEEpeerreviewmaketitle

\newpage

\section{Introduction}
The image classification issue has always been one of the most important problems in the field of the computer vision. Many researchers have focused on achieving improvements in prediction accuracy of image classification. On this long and challenging road, many important branches have been derived out and several worthy research topics have been provided. Since neural networks were firstly proposed in the twentieth century, scientists have inestigated this fundamental problem for plenty of years. However, the early neural network models have not been applied on large-scale applications due to the limitations of computational power and insufficient data sets. The proposal of back-propagation algorithm \cite{lecun1989backpropagation} and the rapid increase of computational power in recent years have made it possible to train neural networks, especially deep neural networks. With more layers of the neural network models designed, the structure is becoming more complex and performance of the model is becoming better. 

A lot of researches \cite{DU2022115, krizhevsky2012imagenet, paoletti2018deep, real2019regularized, schmidhuber2015overview, simard2003best, simonyan2014very, szegedy2017inception, xie2017aggregated, zoph2018learning} are devoted to improving the accuracy rate of image classification, such as Convolutional Neural Network (CNN) \cite{liu2017survey, zhang2019recent}, Vision Transformer \cite{vaswani2017attention} and their variants. The LeNet-5 was proposed to solve the problem of handwritten digit classification, which was an early application of Convolutional Neural Networks in the field of image recognition. In recent years, the application of attention mechanisms \cite{niu2021review} has further improved the performance of models, in the field of not only Computer Vision \cite{guo2016deep} but also Natural Language Processing \cite{lateef2019survey}. By stacking attention modules, Vision Transformer contains more model parameters and further improves the ability to acquire and preserve information. However, in practice, the number of parameters and complexity of the model are often limited by factors such as memory, CPU or GPU, etc.

Different from the relatively idealized conditions in the laboratory, algorithm scientists often face a variety of problems such as data hungry, model complexity limitation and constraint by computational power in practical application \cite{wu2020recent}. First of all, many practical scenarios are unlikely to provide enough data for neural networks to learn, which leads to the problems of data hungry. Secondly, the limited running memory and storage space do not support the stacking of neural network layers and the increasing in the number of parameters. Meanwhile, the data often changes in real time and the models need to output results as soon as possible, so there is no time to manually adjust the hyper-parameters. Last but not least, due to the impact of insufficient computational power, such as GPU resources, it is impossible to have almost infinite iteration time as in the laboratory. The Adaptive Data Augmentation Framework proposed in this paper solves the above problems. On the premise of increasing the number of parameters by less than one thousandth and keeping the hyper-parameters unchanged, this data augmentation framework not only improves the prediction accuracy of image classification but also reduces the number of iterations required to achieve the highest prediction accuracy.

Data hungry is a problem which is often encountered in practical tasks. Many domains, such as science and medicine, would hardly have datasets the size of ImageNet. The size of dataset limits the ability of model to obtain information from the training set, which affects the final performance of the prediction results. In response to the problem of data hungry, many people have proposed data augmentation algorithms. Among them, the direct operations on an image include flip, rotation, scaling, cropping and shifting. In addition, adding Gaussian noise can also achieve the effect of data augmentation. However, these data augmentation techniques prohibit dynamic adjustment of the parameters during the training process. Thus the effect of the whole augmentation technique has been determined at the beginning of iterations. Therefore, adaptive trainable data augmentation algorithms become a necessary research topic \cite{cubuk2019autoaugment}. It is worth mentioning that the Generative Adversarial Network (GAN) \cite{antoniou2017data, creswell2018generative,  goodfellow2014generative} is a very delicately designed neural network model. GAN enables one network to generate samples that are different from the training set and hard to distinguish from the other network, and uses two subneural networks to compete against each other during the training process to reach a Nash equilibrium. In the process of training, the prediction accuracy is improved, and it can also be regarded as an adaptive data augmentation framework.

In this paper, our idea mainly comes from exploring the complementarity of tensor operators and deep neural networks in the ability of image information extraction. We notice that neural networks built using only tensor operators \cite{newman2018stable} do not reach the SOTA (State of the Art) level. We think that the reason may lie in the insufficient ability of tensor operators to extract local features. Meanwhile, the traditional deep learning network has a certain space for improvement in the extraction of global image information, such as the relative positions of pixels and the potential correlation between pixels \cite{novikov2015tensorizing}. As an RGB image, a digitized image should be considered as a third-order tensor. It is therefore natural to ask whether it is feasible to treat the image as a three-dimensional whole and add it to the operation of the neural network model. There have been several papers proposing tensor deep learning models or tensor high-dimensional functions and achieving some results before us \cite{janzamin2015beating, miao2020generalized, miao2021t,newman2018stable, schutt2017quantum}. However, the complexity of tensor computation and the ability of pure tensor neural networks to capture local features are two important constrains. Fortunately, the T-product operator is introduced to deal with problems related to third-order tensors. Before that, operators could cause a lot of problems with the curse of dimensionality or too much complexity, such as Einstein product and outer product. 

We combine T-product and original learning model for better properties and higher accuracy, where the T-product operator obtains the macroscopic features of the whole, and the original model obtains the microscopic features of the detailed parts. Different from the existing Tensor Network or Tensor Neural Network \cite{bao2016tensor,evenbly2011tensor,evenbly2015tensor,montangero2018introduction}, we propose an \textbf{A}daptive \textbf{D}ata \textbf{A}ugmentation \textbf{F}ramework based on the tensor \textbf{T-}product (\textbf{T-ADAF}). With only increasing the number of parameters by less than 0.1\%, the augmentation of the three-dimensional tensor data of the image is achieved by the T-product operator in a structure-preserving manner, giving three times the amount of data for the original model to be enhanced. Since the deep learning model is not completely discarded, the T-product operator and the model are tightly coupled to obtain richer features of image data, not only global but also local features, not only in spatial domain but also in frequency domain. The numerical experiments will demonstrate the effectiveness of our framework. 

The innovation and contribution of this paper can be summarized as follows.
\begin{itemize}
\item[(a)] This paper proposes an novel adaptive data augmentation framework to solve the problem of data hunger, extracting more fundamental features.
\item[(b)] This paper explores the application of tensor T-product operator in image classification tasks and obtains the SOTA results. According to the references we collect, few experimental results have been disclosed in the literature.
\item[(c)] The prediction accuracy is improved on the premise that the number of model parameters is rarely increasing. This has great practical and theoretical value for many miniaturized deep learning tasks, with storage or memory limited.
\end{itemize}
 
 Moreover, the advantages of the adaptive data augmentation framework proposed in this paper can be listed here. 
\begin{itemize}
\item [(a)]

 \textbf{User Friendliness}. All of image classification networks, $\mathcal F$, can be considered as a function space, $\mathscr F=\{\mathcal F\}$, and T-ADAF can be considered as a functional, $\Theta$, on the network function space, $\mathscr F$. Any model can be embedded in this framework, with all hyperparameters of the original model unchanged. At the same time, the framework only increases less than one thousandth of the original model parameters. This provides great convenience for the use of the framework. 
\item[(b)]
\textbf{Better Performance.} The framework adds the T-product tensor operator on the premise of retaining the advantages of the original deep learning network. This framework can provide three times the amount of data for the model, while improving the model's performances for image data. Meanwhile, T-ADAF is adaptive and the tensor parameters in framework are designed as learnable parameters during training process to be more suitable for classification models.
\item[(c)]
\textbf{More Complete Information Extraction}. In this framework, the original model processes signals in the spatial domain and concentrates more on local features, while the T-Product module processes signals in the frequency domain and pays more attention on global features. Scientists can adjust the weights to determine relative importances on T-Product module and original model.

\end{itemize}

This paper is organized as follows. In Section 2, some fundamental definitions, properties and notations of T-product operator are listed for convenience. Besides, some main results are reviewed. Adaptive Data Augmentation Framework is gained in Section 3. Numerical experiments are designed to compare the model to be promoted with our Adaptive Data Augmentation Framework respectively in Section 4. Finally, in Section 5, we analyze the results and indicate the advantages of Adaptive Data Augmentation Framework.

\section{Tensor T-product}
In this section, tensor T-product from the numerical linear algebra will be introduced. At the same time, some essential theorems and properties will be analyzed, which will be utilized for the following sections \cite{hao2013facial, kilmer2011factorization, kilmer2013third, newman2018stable, Che2022efficient, Chen2022tensor}.
\subsection{Notations and Index}
Suppose that we have a tensor $\mathcal A\in\mathbb R^{m\times n\times p}$. Its frontal, horizontal and lateral slices are  defined respectively
\begin{equation}
\begin{cases}
&\mathcal A[:,\ :,\ i]\in\mathbb R^{m\times n},\ \ i=1,2,\ldots,p\\
&\mathcal A[j,\ :,\ :]\in\mathbb R^{n\times p},\ \ j=1,2,\ldots,m\\
&\mathcal A[:,\ k,\ :]\in\mathbb R^{m\times p},\ \ k=1,2,\ldots,n.
\end{cases}
\end{equation}
This can be illuminated more clearly by the following figures.

\begin{figure}[ht]
\centering
\includegraphics[width=2.5in]{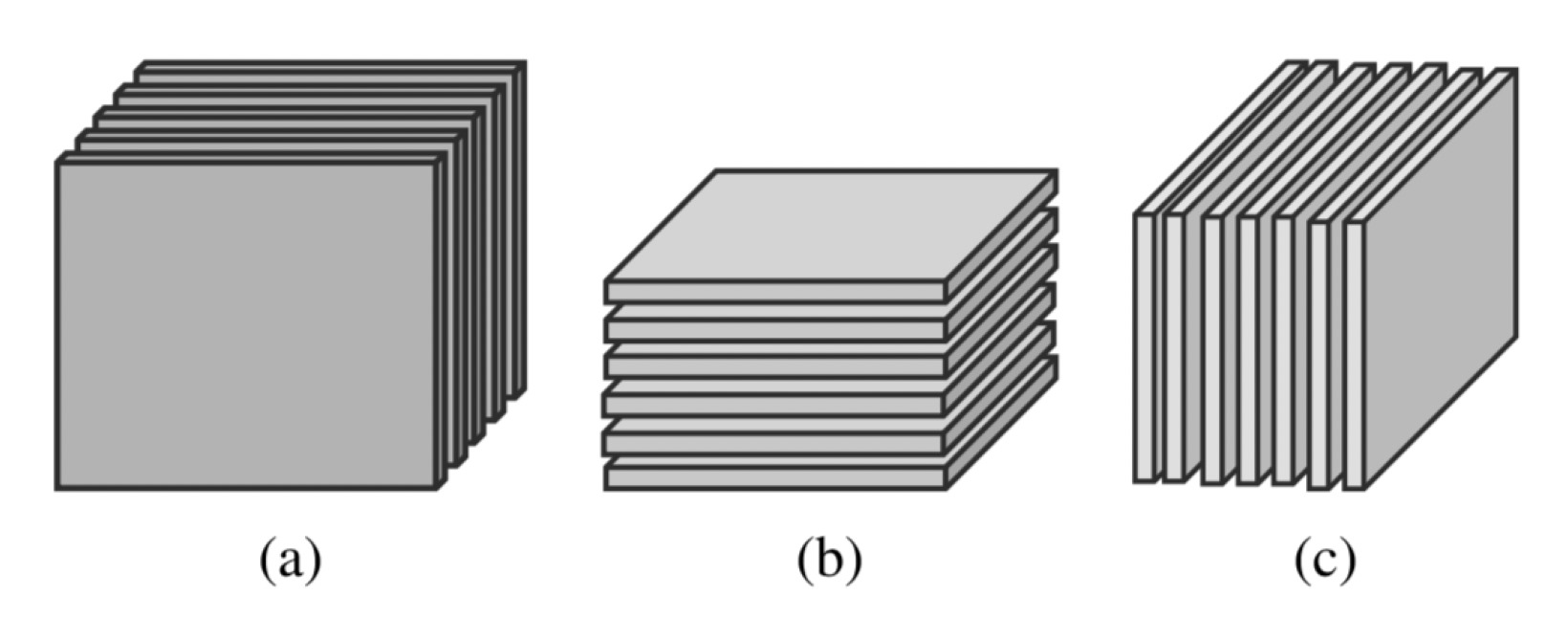}
\caption{(a) frontal, (b) horizontal, (c) lateral slices of a three order tensor}
\end{figure}
The frontal slices can be denoted as $A^{(i)}\in\mathbb R^{m\times n}$, for convenience. Furthermore, we also define some tensor operators, bcirc, unfold and fold, respectively,
\begin{equation*}
{\rm bcirc}(\mathcal A)=
\begin{bmatrix}
A^{(1)} & A^{(p)} & A^{(p-1)} & \cdots & A^{(2)}\\
A^{(2)} & A^{(1)} & A^{(p)} & \cdots & A^{(3)}\\
A^{(3)} & A^{(2)} & A^{(1)} & \cdots & A^{(4)}\\
\vdots & \ddots & \ddots & \ddots & \vdots\\
A^{(p)} & A^{(p-1)} & A^{(p-2)} & A^{(2)} & A^{(1)}
\end{bmatrix}\in\mathbb{R}^{mp\times np}, 
\end{equation*}
\begin{equation*}
{\rm unfold}(\mathcal A):=\begin{bmatrix}
 A^{(1)}\\
 A^{(2)}\\
\vdots\\
 A^{(p)}\\
\end{bmatrix}\in\mathbb{R}^{mp\times n}, \quad
{\rm fold}({\rm unfold}(\mathcal A))=\mathcal A.
\end{equation*}
The corresponding inverse operation of bcirc,
$$
{\rm bcirc}^{-1}:\mathbb R^{mp \times np}\rightarrow\mathbb R^{m\times n\times p}
$$
 is defined as 
 $$
 {\rm bcirc}^{-1}({\rm bcirc}(\mathcal A))=\mathcal A.
 $$
 \subsection{Tensor T-product}

 \begin{definition}{\rm  (Tensor T-product)}
 
 Suppose that there are $\mathcal A\in \mathbb{R}^{m\times n\times p}$ and $\mathcal B\in \mathbb{R}^{n\times s\times p}$. The T-product between $\mathcal A$ and $\mathcal B$ is an $m\times s\times p$ tensor defined by
 \begin{equation*}
 \mathcal A*\mathcal B:={\rm fold}({\rm bcirc}(\mathcal A){\rm unfold}(\mathcal B)).
 \end{equation*}
 \end{definition}
 % The tensor t-function: A definition for functions of thirdorder tensors
 From now on, we will use `$*$’ as T-product in the rest of this paper \cite{newman2018stable}.
 
\begin{definition}{\rm (Transpose and Conjugate Transpose) }

If $\mathcal A$ is a third order tensor, whose size is $m\times n \times p$, then the transpose $\mathcal A^\top$ could be defined from transforming all of the frontal slices and reversing the order of the transposed frontal slices from 2 to $p$. Similarly, the conjugate transpose $\mathcal A^H$ could also be defined from conjugating all of the frontal slices and reversing the order of the transposed frontal slices from 2 to $p$. Writing these two relationships in MATLAB mathematical forms, we have
 \begin{equation}
 \begin{cases}
 \mathcal A^\top[:, :, 0] &=  \mathcal A[:, :, 0]^\top,\\
 \mathcal A^\top[:, :, i] &=  \mathcal A[:, :, p-i]^\top\ \ {\rm for} \ i=1,2,\ldots,p-1,\\
 \mathcal A^H[:, :, 0] &=  \mathcal A[:, :, 0]^H,\\
  \mathcal A^H[:, :, i] &=  \mathcal A[:, :, p-i]^H\ \ {\rm for} \ i=1,2,\ldots,p-1.
 \end{cases} 
 \end{equation}
\end{definition} 
\begin{definition}
{\rm (Identity Tensor)} The $n\times n\times p$ identity tensor $\mathcal I_{nnp}$ is defined as a tensor whose first frontal slice is the $n \times n$ identity matrix, and whose other frontal slices are all zeros.
\end{definition}

It is easy to check that for all $\mathcal A\in\mathbb{R}^{m\times n\times p}$, there exists $\mathcal A*\mathcal I_{nnp}=\mathcal I_{mmp} * \mathcal A=\mathcal A$. 

%\begin{definition}
%{\rm (Tensor T-Inverse)}
%The inverse tensor of frontal square tensor $\mathcal A\in\mathbb{R}^{n\times n\times p} $ could be defined as $\mathcal A^{-1}$, which satisfies 
%\begin{equation*}
%\mathcal A^{-1} * \mathcal A=\mathcal I_{nnp},\ \ \mathcal A * \mathcal A^{-1}=\mathcal I_{nnp}.
%\end{equation*}
%\end{definition}

 \begin{lemma}
 \label{TTlemma}{\rm \cite{lund2020tensor}}
  Using the above definitions of T-product and bcirc operator, there exists
  
{\rm (1)} $ bcirc(\mathcal A*\mathcal B) = bcirc(\mathcal A)\cdot bcirc(\mathcal B),$ \ {\rm (2)} $(\mathcal A * \mathcal B)^H=\mathcal B^H * \mathcal A^H$, $(\mathcal A * \mathcal B)^\top=\mathcal B^\top * \mathcal A^\top$,

{\rm (3)} $bcirc(\mathcal A^\top)=bcirc(\mathcal A)^\top$,\  $bcirc(\mathcal A^H)=bcirc(\mathcal A)^H$.
 \end{lemma}
 It can be easy to notice that the complexity of matrix multiplication to compute T-product by definitions increases rapidly as the dimensions of these tensors goes up, that is
 $
 O(mp\times np\times s)=O(mnsp^2).
 $
 Fortunately, T-product can be accelerated by the Discrete Fourier Transform (DFT) and the Fast Fourier Transform (FFT). Let $F_n$ be the $n\times n$ DFT matrix,
$$
F_n=\frac{1}{\sqrt{n}}
\begin{bmatrix}
1 & 1 &1 & \cdots & 1 \\
1 & \omega^{1}& \omega^{2}& \cdots & \omega^{n-1} \\
1 & \omega^{2}& \omega^{4}& \cdots & \omega^{2(n-1)} \\
\vdots& \vdots& \vdots&\ddots&\vdots\\
1 & \omega^{n-1}& \omega^{2(n-1)}& \cdots & \omega^{(n-1)(n-1)} \\
\end{bmatrix},
$$
where 
$
\omega=e^{-\frac{2\pi i}{n}}
$
is the primitive $n$-th root of unity in which $i^2=-1$.  It is well-known that a block circulant
matrix can be block diagonalized by using the Fourier transform \cite{jin2003developments},
\begin{equation}
\label{bcirc_ope}
\begin{aligned}
{\rm bcirc}(\mathcal A)=\left(F^H_p\otimes I_m\right)
\begin{bmatrix}
A_1 & & & \\
 & A_2 & & \\
 & & \ddots & \\
 & & & A_p
\end{bmatrix}
\left(F_p\otimes I_n\right),
\end{aligned}
\end{equation}
where $F_p^H$ is the conjugate transpose of $F_p$ and ‘$\otimes$’ is the Kronecker product \cite{horn2012matrix}.
\subsection{FFT Algorithm to Compute T-product}

Following the analysis of Section 2.2, we can generate the algorithm to compute T-product based on the FFT method as follows. By recursively dividing the array into even and odd indexed elements, referring to Subsection 1.4 of \cite{golub2013matrix}, the FFT can reduce computational complexity of the DFT. Here we will introduce the well-known FFT algorithm, where $
\omega=e^{-\frac{2\pi i}{n}}
$
is the primitive $n$-th root of unity and $x(s,t,n)=(x(s), x(s+t), x(s+2t), \ldots, x(s+rt))$ with $x(s+rt) < n \leq x(s+(r+1)t)$.

 \begin{breakablealgorithm}
 	\caption{ Fast Fourier Transform Algorithm }
 	\label{alg:Framwork}
 	\begin{algorithmic}[0] %这个1 表示每一行都显示数字
 	\Require ~ $x\in\mathbb{R}^{n}$ and $n=2^t$
 	
 	\Ensure ~ the Discrete Fourier Transform of $x$, $\hat x = \text{fft}(x, n)$
 	
 	\textbf{If} $n=1$:
 	
 	\ \ \ \ $\hat x = x$
 	
 	\textbf{Else} 
 	
  	\ \ \ \ $m = n / 2$
  	
  	\ \ \ \ $d= (1, \omega, \ldots, \omega^{m-1})^\top$
  	
  	\ \ \ \ $\hat x = \left[\begin{matrix}
  	\text{fft}(x(1:2:n), m) + d.*\text{fft}(x(2:2:n), m)\\
  	\text{fft}(x(1:2:n), m) - d.*\text{fft}(x(2:2:n), m)
  	\end{matrix}
  	\right]$
  	
 	\textbf{End} 
 	
 	\end{algorithmic}
 \end{breakablealgorithm}
%Using notations from PyTorch \cite{paszke2017automatic}, a deep-learning library which provides probability to compute the differentiation, we define 
%$$
%\widehat{\mathcal A}=\mbox{torch.fft.fft}(\mathcal A,\ {\rm dim}=2)
%$$
%as the Fast Fourier Transform operation on $\mathcal A$. While in MATLAB, this command should be changed into 
%$$\widehat{\mathcal A}=\mbox{fft}(\mathcal A,[\ ],3).$$
%Symmetrically, the inverse Fast Fourier Transform (IFFT) can be defined. 

In this algorithm, ``.*" denotes element-wise multiplication between two vectors. Similarly, inverse Fast Fourier Transform (iFFT) operation can be defined. In \cite{kilmer2011factorization, kilmer2013third}, Kilmer has shown that (\ref{bcirc_ope}) could be equal to take the DFT and iDFT operation on $\mathcal A$ along its third dimension. For $i=1,\ldots,m$ and $j=1,\dots,n$, the tube element vector of $\mathcal A\in \mathbb{R}^{m\times n\times p}$ is noted as $\mathcal A[i,j,:]$. 
 
\begin{breakablealgorithm}
	\caption{ T-product Algorithm based on FFT Method}
	\label{alg:Framwork}
	\begin{algorithmic}[0] %这个1 表示每一行都显示数字
	\Require ~ $\mathcal A\in\mathbb R^{m\times n\times p}$ and $\mathcal B \in\mathbb R^{n\times s\times p}$
	
	\Ensure ~ $\mathcal C = \mathcal A*\mathcal B\in\mathbb R^{m\times s\times p}$\\
	\textbf{Step 1:} Take FFT operation on $\mathcal A$
	
	\textbf{For} $i=1,\ldots,m$ and $j=1,\dots,n$,
	
	$\ \ \ \ \widehat{\mathcal A}[i,j,:]$ = fft($\mathcal A[i,j,:],$ $p$)
	
	\textbf{End}\\
	\textbf{Step 2:} Take FFT operation on $\mathcal B$
	
	\textbf{For} $j=1,\ldots,n$ and $t=1,\dots,s$,
		
		$\ \ \ \ \widehat{\mathcal B}[j,t,:]$ = fft($\mathcal A[j,t,:],$ $p$)
		
		\textbf{End}\\
	\textbf{Step 3:} Compute the frontal piecewise product between $\mathcal A$ and $\mathcal B$
 	
	\textbf{For} $k=1,\cdots,p$,
	
			$\ \ \ \ \widehat{\mathcal C}[:, :, k]$ = $\widehat{\mathcal A}[:, :, k] \cdot \widehat{\mathcal B}[:, :, k]$
			
	\textbf{End}\\
		\textbf{Step 4:} Take iFFT operation on $\mathcal C$
	
	\textbf{For} $i=1,\ldots,m$ and $t=1,\dots,s$,
			
			$\ \ \ \ {\mathcal C}[i,t,:]$ = ifft($\widehat{\mathcal C}[i,t,:],$ $p$)
			
			\textbf{End}

	\end{algorithmic}
\end{breakablealgorithm}

\begin{remark}

In this algorithm, the computation complexity is
$$
O(m\times n\times s\times p) = O(mnsp).
$$
Meanwhile, the cost of FFT in one tube is $O(p\log p)$. So the whole cost of T-product based on FFT method between $\mathcal A\in\mathbb R^{m\times n\times p}$ and $\mathcal B\in\mathbb R^{n\times s\times p}$ is
$$
O(mnsp+mnp\log p+nsp\log p),
$$
which is efficient than just using definition of T-product.
\end{remark}
%\subsection{Comparison with other Tensor Operator}
%The Einstein product performs on tensor $\mathcal A:=(a_{i_1,\cdots,i_M,j_1,\cdots,j_N})\in\mathbb C^{I_1\times\cdots\times I_M\times J_1\times \cdots\times J_N}$ and $\mathcal B:=(b_{j_1,\cdots,j_N,k_1,\cdots,k_L})\in\mathbb C^{I_1\times\cdots\times I_M\times K_1\times \cdots\times K_L}$. 
%
%Then Einstein Product
%$\mathcal A*_N\mathcal B\in\mathbb C^{J_1\times\cdots\times J_N\times K_1\times \cdots\times K_L}$ is given by
%\begin{equation}
%\left(\mathcal A *_N\mathcal B\right)_{i_1,\cdots,i_M,k_1,\cdots,k_L}=
%\end{equation}
\subsection{Reviews of Tensor Deep Learning Methods}

As a high-order data storage structure, tensor has the property of preserving structures and connections among different data points. On contrary, vectorization always destroys the data structure, which might limit the performance of model. In plenty of deep learning tasks, such as color images, video, and a variety of time-varying signals, tensors have obvious advantages. At the same time, the nonlinearity and flexibility implied by the tensor operator offer more freedom to the model construction. There has been a lot of research on tensor deep learning to propose many important ideas and models. For instance, in \cite{shi2022fuzzy}, Shi, Ma, and Fu propose a fuzzy support tensor product method for image classification, utilizing lots of unlabeled image data and improving the network robustness. Meanwhile, in \cite{naumov2023tetra}, Naumov {\it et al.} uses Tetra-AML toolbox and tensor network to provide model compression for deep network like ResNet-18. In addition to some traditional senses, the tensor network also makes improvement in medical image classification \cite{selvan2020tensor} and quantum-based image recognition \cite{liu2019machine, orus2019tensor}.  Apart from RGB images, tensor neural networks have many applications in hyperspectral image (HSI) image recognition. A tensor-based dictionary self-taught learning classification method is proposed for solving precise object classification problems based on Hyperspectral imagery with limited training data \cite{liu2022tensor1}. Besides, in \cite{LIU2021215}, Liu, Ma, and Wang studied HSI and intuitively represented HSI as a third-order tensor. In the sparse tensor dictionary learning algorithm, the joint spatial-spectral information can significantly improve the accuracy of HSI classification, combining with tensor Turker decomposition \cite{Che2019randomized, Che2022perturbation}. Meanwhile, different from the traditional vector-based or matrix-based methods, in \cite{LIU2020107361}, tensor technique is adopted to extract the joint spatial-spectral tensor features. Tensor T-product \cite{newman2018stable} is also studied to model the forward propagation in neural networks. Besides, some other researches focus on tensor deep learning modeling \cite{pellionisz1985tensor,singh2010tensor,singh2011tensor,socher2013reasoning}.
\section{Adaptive Data Augmentation Framework}
In this section, the Adaptive Data Augmentation Framework based on T-product will be derived and the Tensor Back-Propagation algorithm will be introduced. 
\subsection{Adaptive Data Augmentation Framework based on Tensor T-product}
In such a $K$-classification image recognition task, there is a model, $\mathcal F$, that needs to be improved. Assume that the parameter space of this model is $\Phi$. The $i$-th image data is considered in RGB (Red, Green, Blue) as a three-order tensor $\mathcal A^{(i)}\in\mathbb{R}^{m\times n\times 3}$. 

Following this, the label of observed data $\mathcal A^{(i)}$ which belongs to $k$-th class is a $k$-dimension one-hot vector $\bm y_i$, where
$$
y_{ij}=
\begin{cases}
1, \ \ \ j=k,\\
0, \ \ \ j \neq k.
\end{cases}
$$

Then our goal is to make sure that
\begin{equation}
k=\arg\max\mathcal F(\mathcal A^{(i)};\Phi),
\end{equation}
where $\mathcal F(\mathcal A^{(i)};\Phi)$ is also a $k$-dimension prediction vector,
$
p_i=\mathcal F(\mathcal A^{(i)};\Phi).
$

There are many kinds of loss function that we could choose. In this paper, what we utilize is Cross Entropy function,
\begin{equation}
\mathcal L=-\frac{1}{N}\sum_{i}\sum_{k=1}^Ky_{ik}\log(p_{ik}),
\end{equation}
where $y_{ik}$ is the label of data $i$ and $p_{ik}$ is the predicted probability that the observed sample $i$ belongs to the class $k$.

As we introduce about the T-product, if we permute a image data $\mathcal A_0\in \mathbb R^{m\times n\times 3}$ into ${\mathcal A}_1\in \mathbb R^{m\times 3\times n}$ and ${\mathcal A}_2\in \mathbb R^{n\times 3\times m}$, the parameter tensors, $\mathcal W_1 \in \mathbb R^{3\times 3\times n}$ and $\mathcal W_2 \in \mathbb R^{3\times 3\times m}$, could be multiplied to maintain the size of the permuted image tensor data. The sizes of
$$
\mathcal A_1 * \mathcal W_1 \in \mathbb R^{m\times 3\times n}
$$
and 
$$
\mathcal A_2 * \mathcal W_2 \in \mathbb R^{n\times 3\times m}
$$
are both similar to those of $\mathcal A_1$ and $\mathcal A_2$. The following figure can help readers understand the process better.

\begin{figure}[ht]
\centering
\includegraphics[width=4.7in]{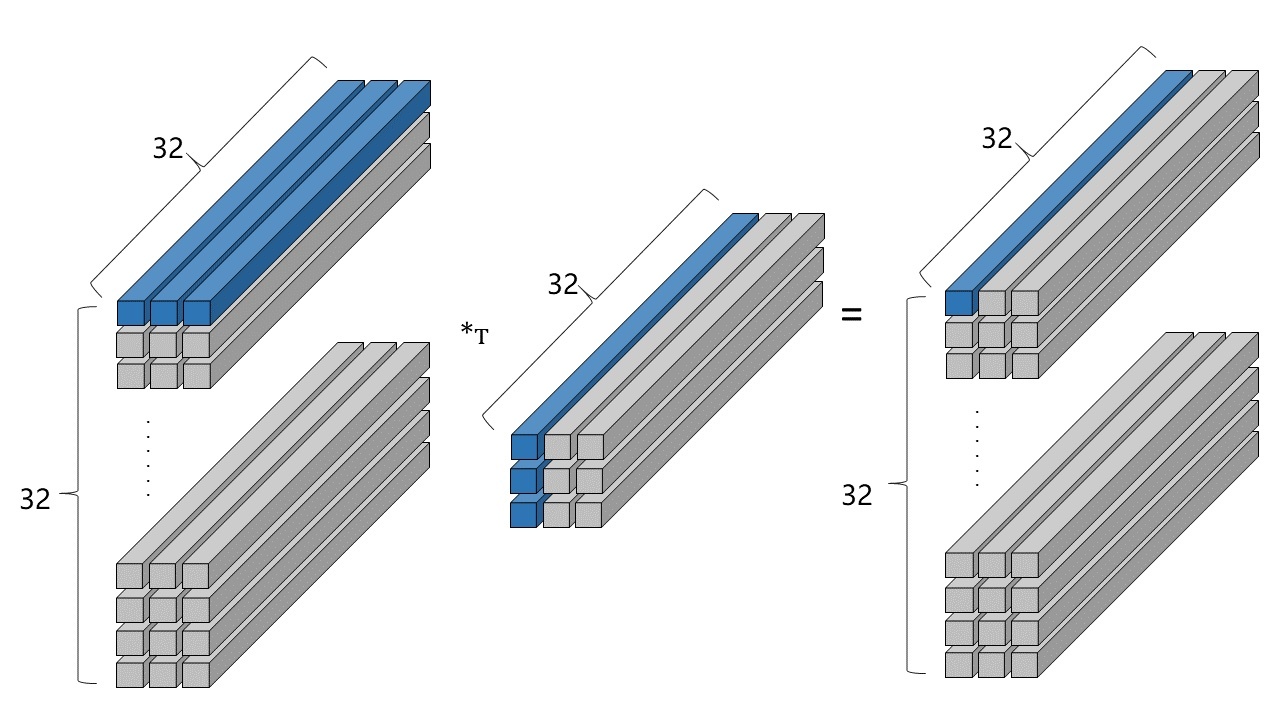}
\caption{Tensor T-Product in T-ADAF between $\mathcal A_i\in\mathbb{R}^{32\times 3\times 32}$ and $\mathcal W_i\in\mathbb{R}^{3\times 3\times 32},\ i=1,2$}
\end{figure}

The blue tubes in the above Figure shows how FFT and dot product work in T-product. All of the data can be permuted into a normal image size $\mathbb R^{m\times n\times 3}$. As a consequence, they can be treated as two new images and taken into the processing of classification. From the model, $\mathcal F$, three results can be gained, $\bm r=(r_0,r_1,r_2)^\top$,
\begin{equation}
\begin{cases}
r_0&=\mathcal F(\mathcal A_0;\Phi),\\
 r_1&=\mathcal F(\mathcal A_1;\Phi),\\
r_2&=\mathcal F(\mathcal A_2;\Phi).
\end{cases}
\end{equation}
These three results are combined together to gain the final prediction result
\begin{equation}
r=\bm w^\top \bm r=w_0r_0+w_1r_1+w_2r_2,
\end{equation}
where $\bm w=(w_0, w_1, w_2)^\top$ is the corresponding weight of each results.
\begin{remark}
It should be noticed that:
\begin{itemize}
\item[\rm (1)] T-product could be considered as the projections into different dimensions and the weights show the importance of these corresponding projections.
\item[\rm (2)] The weights $w_i$, $i=0,1,2$ could be chosen as any vectors, which satisfy
\begin{equation}
w_0+w_1+w_2=1,
\end{equation}
 such as $\left(\frac{1}{3},\frac{1}{3},\frac{1}{3}\right)$,  $\left(\frac{1}{2},\frac{1}{4},\frac{1}{4}\right)$ or $(\frac{2}{5}, \frac{3}{10}, \frac{3}{10})$.
%\item[3)] From (2) of \textbf{Lemma \ref{TTlemma}} and other properties of T-transposing, the learning processing of parameters of $\mathcal A_i$ and $\mathcal A_i^\top$ is essentially equivalent, i.e.,
%\begin{equation}
%\mathcal A_i^\top * \mathcal W=\left(\mathcal W^\top * \mathcal A_i\right)^\top.
%\end{equation}
\item[\rm (3)] The weights in T-product, $\mathcal W_i,\ i=1,2$ are all initialized by identity tensor on T-product, $\mathcal I_{nnp},$ to make sure that 
$$
\mathcal F(\mathcal A_0;\Phi)=\mathcal F(\mathcal A_1;\Phi)=\mathcal F(\mathcal A_2;\Phi).
$$
Moreover, considering T-ADAF as a functional on the model function space $\mathscr F=\{\mathcal F\}$ and noting it as $\Theta$, there exists that
$$
\Theta(\mathcal F|\Phi^{(0)},\mathcal W_1^{(0)},\mathcal W_2^{(0)})=\mathcal F(\ \cdot\ |\Phi^{(0)}), \ \ for\   \mathcal F\in\mathscr F,
$$
where $\Phi^{(0)},\mathcal W_1^{(0)},\mathcal W_2^{(0)}$ are the initial parameters of original model and T-product, respectively. 

\end{itemize}
\end{remark}

\begin{figure}[htbp]
\begin{tikzpicture}
% \begin{figure}[htb]
\label{liuchengtu}
		\centering
		\scriptsize  
		\tikzstyle{yuan} = [rectangle, rounded corners, minimum width = 2cm, minimum height=8mm,text centered, draw = black, fill=green]
		\tikzstyle{sanjiao} = [trapezium, trapezium left angle = 70,trapezium right angle=110,minimum height=1cm,text centered,draw=black,fill=blue!30]
		\tikzstyle{io} = [trapezium, trapezium left angle=70, trapezium right angle=110, minimum width=2cm, minimum height=1cm, text centered, draw=black]
		\tikzstyle{process} = [rectangle, minimum width=3cm, minimum height=1cm, text centered, draw=black， fill = yellow]
		\tikzstyle{decision} = [diamond, aspect = 3, text centered, draw=black]
		\tikzstyle{fang}=[rectangle,minimum width = 2cm, minimum height=8mm,draw,thin,fill=yellow]  
				%定义语句块的颜色,形状和边
		\tikzstyle{test}=[diamond,aspect=2,draw,thin]  
				%定义条件块的形状,颜色
		\tikzstyle{point}=[coordinate,on grid]  
		
		%像素点,用于连接转移线
		%[node distance=10mm,auto,>=latex',thin,start chain=going below,every join/.style={norm},] 
		%start chain=going below指明了流程图的默认方向，node distance=8mm则指明了默认的node距离。这些可以在定义node的时候更改，比如说 
		%\node[point,right of=n3,node distance=10mm] (p0){};  
		%这里声明了node p0，它在node n3 的右边，距离是10mm。
		\node[yuan] (start) at (-1.5, 6){Image, $\mathcal P\in\mathbb R^{m\times n\times 3}$};
		
		\node[fang] (O0) at (-3.6, 4.5) {Identity};
		\node[fang] (O1) at (0.6, 4.5) {Permute};

		\node[yuan] (P0) at (-4.8, 3) {$\mathcal P_0\in\mathbb R^{m\times n\times 3}$};
		\node[yuan] (P1) at (-1.2, 3){$\mathcal P_1\in\mathbb R^{n\times 3\times m}$};
		\node[yuan] (P2) at (2.4, 3){$\mathcal P_2\in\mathbb R^{m\times 3\times n}$};
		
		\node[fang] (O2) at (-4.8, 1.5) {Identity, $\mathcal I\in\mathbb{R}^{n\times n\times 3}$};
		\node[fang] (O4) at (-1.2, 1.5) {T-product, $\mathcal T_1\in\mathbb R^{3\times 3\times m}$};
		\node[fang] (O5) at (2.4, 1.5) {T-product, $\mathcal T_2\in\mathbb R^{3\times 3\times n}$};
		
		\node[yuan] (P0-2) at (-4.8, 0) {$\mathcal P_0\in\mathbb R^{m\times n\times 3}$};
		\node[yuan] (P1-2) at (-1.2, 0){$\widetilde{\mathcal P}_1\in\mathbb R^{n\times 3\times m}$};
		\node[yuan] (P2-2) at (2.4, 0){$\widetilde{\mathcal P}_2\in\mathbb R^{m\times 3\times n}$};
		
		\node[fang] (O6) at (-3.6, -1.5) {Identity};
		\node[fang] (O7) at (0.6, -1.5) {\makecell{Inverse of Permute Operation\\ and Activation Function}};
		
		\node[yuan] (P0-3) at (-4.8, -3) {$\mathcal P_0\in\mathbb R^{m\times n\times 3}$};
		\node[yuan] (P1-3) at (-1.2, -3){$\widehat{\mathcal P}_1\in\mathbb R^{n\times 3\times m}$};
		\node[yuan] (P2-3) at (2.4, -3){$\widehat{\mathcal P}_2\in\mathbb R^{m\times 3\times n}$};
		
		\node[sanjiao] (Model) at (-1.5, -4.8) {\makecell{Initial Model for \\
		Image Classifacation, $\mathcal F$}};
		
		\node[yuan] (P0-4) at (-4.8, -6.6) {Result 1, $r_0$};
		\node[yuan] (P1-4) at (-1.2, -6.6) {Result 2, $r_1$};
		\node[yuan] (P2-4) at (2.4, -6.6) {Result 3, $r_2$};
		
		\node[yuan] (end) at (-1.5, -8.3) {Result for 
				Output, $r$};
		%\node[format] (n0) at(4,4){A}; 直接指定置 
		%定义完node之后进行连线,
		%\draw[->] (n0.south) -- (n1); 带箭头实线
		%\draw[-] (n0.south) -- (n1); 不带箭头实线
		%\draw[<->] (n0.south) -- (n1.north);   双箭头
		%\draw[<-,dashed] (n1.south) -- (n2.north); 带箭头虚线 
		%\draw[<-] (n0.south) to node{Yes} (n1.north);  带字,字在箭头方向右边
		\draw[->] (start.south) to (O0.north);  
		\draw[->] (start.south) to (O1.north);  
		\draw[->] (O0.south) to (P0.north);  
		\draw[->] (O1.south) to (P1.north);  
		\draw[->] (O1.south) to (P2.north);  
		\draw[->] (P0.south) to (O2.north);  
		\draw[->] (P1.south) to (O4.north);  
		\draw[->] (P2.south) to (O5.north);  
		\draw[->] (O2.south) to (P0-2.north);  
		\draw[->] (O4.south) to (P1-2.north);  
		\draw[->] (O5.south) to (P2-2.north);  
		\draw[->] (P0-2.south) to (O6.north);  
		\draw[->] (P1-2.south) to (O7.north);  
		\draw[->] (P2-2.south) to (O7.north);  
		\draw[->] (O6.south) to (P0-3.north);  
		\draw[->] (O7.south) to (P1-3.north);  
		\draw[->] (O7.south) to (P2-3.north);  
		\draw[->] (P0-3.south) to (Model.north);  
		\draw[->] (P1-3.south) to (Model.north);  
		\draw[->] (P2-3.south) to (Model.north);  
		\draw[->] (Model.south) to (P0-4.north);  
		\draw[->] (Model.south) to (P1-4.north);  
		\draw[->] (Model.south) to (P2-4.north);  
		%\draw[->] (n1.north) to[out=60,in=300] node{Yes} (n0.south);  曲线
		%\draw[->,draw=red](n2)--(n1);  带颜色的线
		\draw[->] (P0-4.south) to node{$w_0$} (end.north);  
		\draw[->] (P1-4.south) to node{$w_1$} (end.north);  
		\draw[->] (P2-4.south) to node{$w_2$} (end.north);  
		  
\end{tikzpicture}
\caption{Model Enhancement Framework based on T-product}
\label{liuchengtu}
\end{figure}
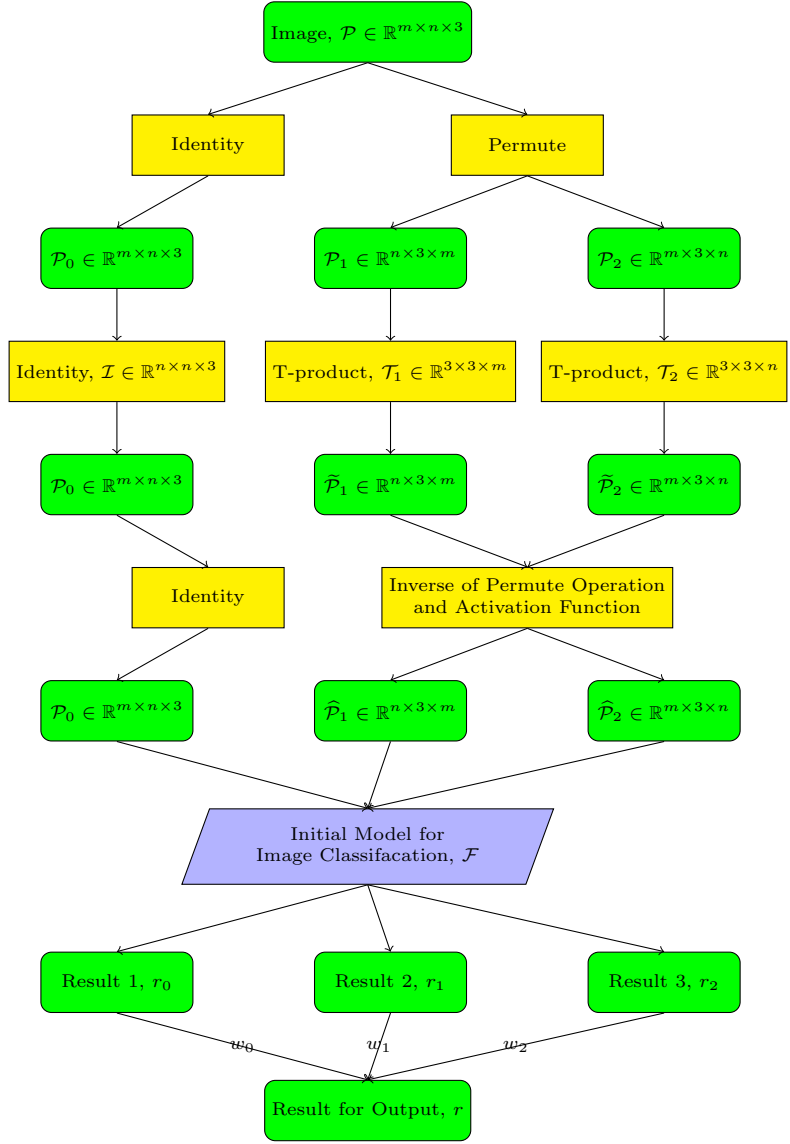

For convenience, Figure \ref{liuchengtu} is the schematic illustration of the framework based on T-product. In the full of framework, all of the hyper-parameters when training are maintained, such as learning rates, batch size, etc. 

\subsection{Tensor Back-Propagation based on T-product}
In this subsection, Tensor Back-Propagation (BP) \cite{nielsen2015neural, rumelhart1986learning} based on T-product will be introduced. If an algorithm engineer could rewrite the BP algorithm in Cuda or Pytorch, it will save some time to train the neural network with our Adaptive Data Augmentation Framework. These conclusion has been derived  by Newman {\it et al.} \cite[Derivation 4.1]{newman2018stable}.
\begin{theorem}
Suppose that forward propagation of tensor T-product operation in a neural network is 
\begin{equation}
\mathcal A_{j+1}=\sigma(\mathcal Z_{j+1})= \sigma(\mathcal W_j *\mathcal A_j+\mathcal B_j),
\end{equation}
for $j=0,1,\ldots,N$, where $\sigma(\cdot)$ is the activation function. $\mathcal M$ is the whole tensor neural network. Then the tensor back-propagation formulas can be computed by,
\begin{equation}
\delta\mathcal A_N=\mathcal W_N^\top*\frac{\partial \mathcal M}{\partial \mathcal A_N},
\end{equation}
\begin{equation}
\delta\mathcal A_j=\mathcal W_j^\top*(\delta\mathcal A_{j+1}\odot \sigma ' (\mathcal Z_{j+1})),\ \ {\rm for}\ \  i=0,1,\ldots,N-1,
\end{equation}
\begin{equation}
\delta\mathcal W_j=(\delta\mathcal A_{j+1}\odot \sigma ' (\mathcal Z_{j+1}))*\mathcal A_j^\top,\ \ {\rm for}\ \  i=0,1,\ldots,N-1,
\end{equation}
where 
\begin{equation*}
\delta \mathcal A_j:=\frac{\partial \mathcal  M}{\partial\mathcal A_j},\ \ {\rm for}\ \  i=0,1,\ldots,N,
\end{equation*}
is the erros on the $j$-th layer, $\delta'$ is the derivative of the activation function and `$\odot$' is the Hadamard element-wise product. 
\end{theorem}

\begin{figure}[htbp]
 \centering
 \includegraphics[width=4in]{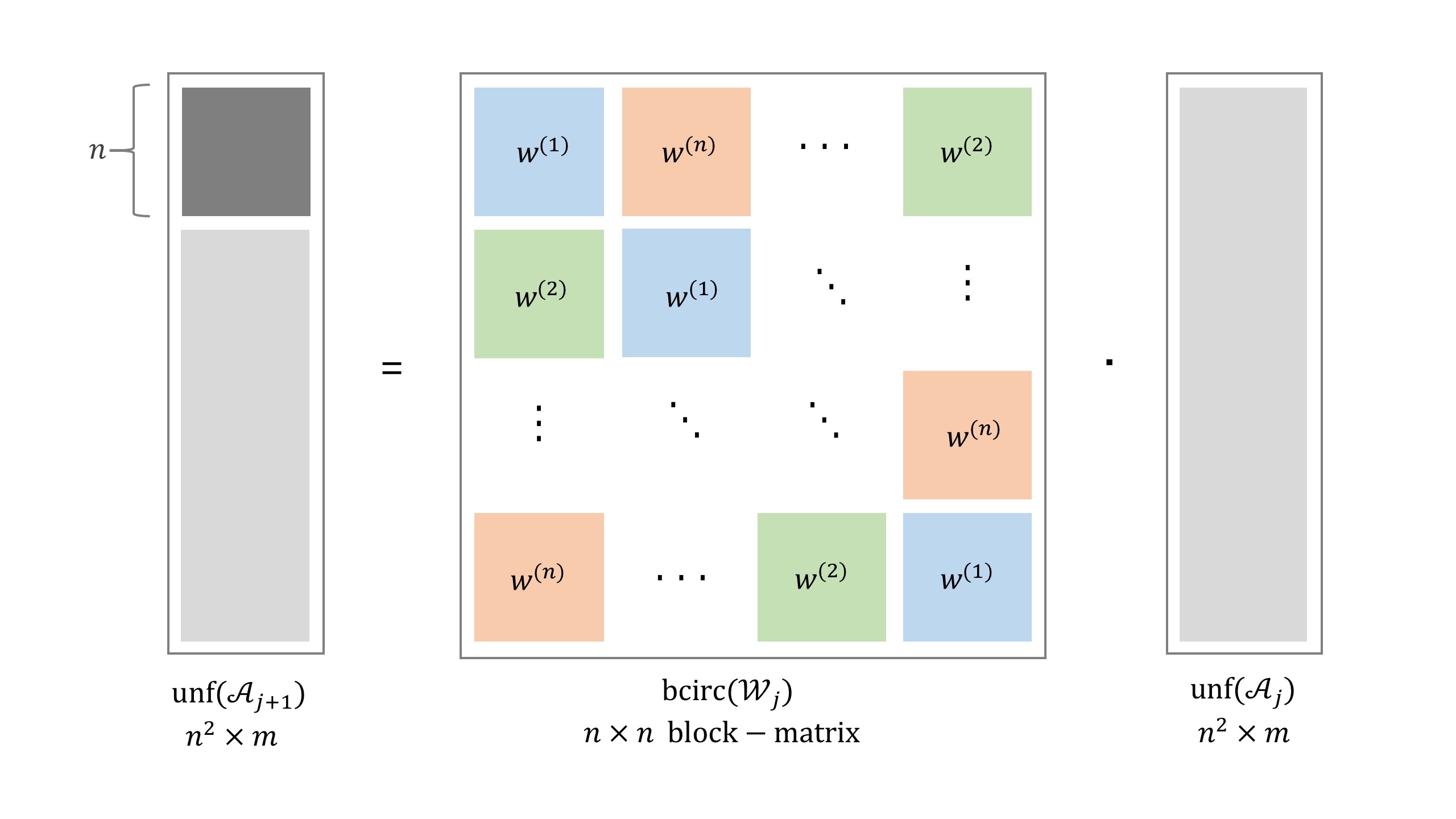}
 \caption{Definition of Tensor T-Product with Bcirc Operator}
 \end{figure}

Suppose that the auto back-propagation operator in PyTorch or TensorFlow is noted as $\Psi(\delta_{final};\mathcal F)$, where $\mathcal F$ is the original model and $\delta_{final}$ is the error in the final step between the actual target vector and final predicted result. From the above definition,
\begin{equation}
\Psi(\delta_{final};\mathcal F)=\delta_T,
\end{equation}
where $\delta_T$ is the error after T-product.

\begin{breakablealgorithm}
	\caption{Back-Propagation Algorithm for T-ADAF}
	\begin{algorithmic}[0] %这个1 表示每一行都显示数字
	\Require ~ 
	
	The final error between  the actual target vector and final predicted result, $\delta_{final}$.
	
	The original model, $\mathcal F$.
	
	{Initial data before T-product, $\mathcal A_0\in\mathbb{R}^{m\times n\times 3}$.}
	\Ensure ~ 
	
	The error of weights in T-product, $\delta\mathcal W_0$.\\
		{\textbf{Hyper-parameters:}}
		
		{The batch size, $B$.}
		
		{The maximum number of iteration epochs, $N$.}
		
		{The learning rate schedule, $\eta_i,\ i=1, \ldots,N.$}
		
		{The optimizer to train the model parameters, $\mathcal O$. }
		
		{The weights of each prediction results, $w$.}
		
		{Initial weights in T-product, $\mathcal W_0\in\mathbb{R}^{3\times 3\times n}$.}\\
	\textbf{Step 1:} Compute the error after T-product, $\delta_T$, using the auto back-propagation operator in PyTorch,
	\begin{equation*}
	\Psi(\delta_{final};\mathcal F)=\delta_T.
	\end{equation*}\\
	\textbf{Step 2:} Permute $\delta_T\in\mathbb{R}^{m\times n\times 3}$ into $\widehat\delta_T\in\mathbb{R}^{m\times 3\times n}$ and $\mathcal A_0\in\mathbb{R}^{m\times n\times 3}$ into $\widehat{\mathcal A_0}\in\mathbb{R}^{m\times 3\times n}$.\\
	\textbf{Step 3:} Compute the permuted error of weights in T-product, ${\delta\mathcal W_0}^\top$.\\
	\ \ \ \ \textbf{Step 3.1:} Compute
	\begin{equation*}
	\widehat{\delta\mathcal A_0}^\top=\mathcal W_0*\left(\widehat\delta_T^\top\odot \sigma ' \left(\mathcal W_0^\top*\widehat{\mathcal A_0}^\top\right)\right).
	\end{equation*}
	\ \ \ \ \textbf{Step 3.2:} Compute
	\begin{equation*}
	{\delta\mathcal W_0}^\top=\left(\widehat\delta_T^\top\odot \sigma ' \left(\mathcal W_0^\top*\widehat{\mathcal A_0}^\top\right)\right)*\mathcal A_0.
	\end{equation*}
	\textbf{Step 4:} Permute ${\delta\mathcal W_0}^\top\in\mathbb{R}^{3\times 3\times n}$ into ${\delta\mathcal W_0}\in\mathbb{R}^{3\times 3\times n}$.

	\end{algorithmic}
\end{breakablealgorithm}

\begin{remark}

Compared with the original automatic back propagation algorithm in PyTorch Toolbox, Algorithm 3 does not need to calculate the numerical difference of T-product operator. Therefore, Algorithm 3 avoids multiple computations of the entire frame map and reduces the consumption of calculation.
\end{remark}

\subsection{Analysis of T-ADAF}
According to the definition of T-product, such a circulant convolution-like structure will extract a more comprehensive global data feature information for any side of it. For each frontal slice, the T-product
 
 \begin{equation}
 \mathcal C = \mathcal A * \mathcal  B
 \end{equation}
 could be considered as a circulant convolution-like structure,
 \begin{equation}
 C^{(k)} = A^{(k)}\cdot B^{(1)}+\sum_{i=1}^{k-1}A^{(i)}\cdot B^{(k-i+1)}
 +\sum_{i=k+1}^{n}A^{(i)}\cdot B^{(n-i+k+1)},
 \end{equation}
 for $k=1,2,\ldots,n$.
 
There may be some connections between two blocks of pixels that are far apart and can provide valuable information for model prediction. However, only with the local convolution and pooling operations of the neural network, such features may be diluted or even disappeared. But T-product provides the ability to obtain more global information.

In T-ADAF, T-product plays the role of frequency domain information extraction and analysis. Let us review the algorithm to compute T-product between two tensors,

\begin{equation}
\mathcal A*\mathcal W=iDFT(DFT(\mathcal A) \star DFT(\mathcal W)),
\end{equation}
where `$\star$' means the slice-wise dot product, i.e., 
\begin{equation}
\mathcal A\star \mathcal W[:,\ :,\ i]=\mathcal A[:,\ :,\ i] \ \mathcal W[:,\ :,\ i],
\end{equation}
for $i=1,2,\ldots, N$. DFT transforms image data from spatial domain to frequency domain. Then the linear mapping can be understood as a fully connected layer to extract and analyze the frequency domain signals. Finally, iDFT transfers the data from the frequency domain back to the spatial domain. As a consequence, the data generated through this series of processes are provided to the neural network for further learning as the result of data augmentation, thus improving the performance of original model.

\section{Numerical Results}
% In lots of practical problems, plenty of engineers have found that the model with hyper-parameters which have been adjusted before would not perform well in new cases. Therefore, we divide the whole numerical experiment into two parts, incompletely and completely pre-adjusting hyper-parameters, to avoid the influence of hyper-parameters for our framework. These two parts are depended on whether the hyper-parameters have been pre-adjusting for specific tasks. 

In this section, we aim to make the numerical experiences as close as possible to the actual applications and illuminate that the improvements derive from the T-product-based framework indeed. For these purposes, we present the numerical results in the order by the time when the original models were published. We divide the numerical experiments into two parts, LeNet-5 models and Deep Neural Network models. Looking back on the history of neural networks, LeNet-5 was proposed and studied early. With introduction of deeper models, LeNet-5 gradually faded out of the limelight. However, for many application scenarios, it still has the advantages of smaller memory consumption and fast training speed. For some projects that do not require too high prediction accuracy, small models are a less expensive option. In the past decade, many neural network models with sophisticated design and high performance have been proposed, which we classify as deeper models and explore the effectiveness of this data augmentation framework for these models. We have tested the framework in some undisclosed data sets with some success. In this section, we present the numerical results based on open data sets.
\subsection{Preparations and Hyper-parameter Settings}
The CIFAR-10 dataset consists of $32\times 32\times 3$ RGB images belonging to $10$ different classes, 50,000 training images and 10,000 testing images. Meanwhile, CIFAR-100 dataset consists of $32\times 32\times 3$ RGB images belonging to $100$ different classes, 50,000 training images and 10,000 testing images. Each set of images is equally distributed among different classes. We normalize the (R, G, B) channels of each image with means of $(0.4914, 0.4822, 0.4465)$ and standard deviations of $(0.2023,0.1994,0.2010)$.
		
{The batch size, $B$, is set as 32. The maximum number of iteration epochs, $N$, is 100 for LeNet-5, while $N=200$ for other models. The optimizer to train the model parameters is Adam \cite{kingma2014adam}.  The weights in T-product, $\mathcal W_0\in\mathbb{R}^{3\times 3\times n}$, is initialized as $\mathcal W_0=\mathcal I$, since we want the data augmentation is started from the situation without any changes, i.e., $\mathcal A*\mathcal I=\mathcal A$.} \textcolor{blue}{As for the learning rate schedules, $\eta_i,\ i=1, \ldots,N, $ they are also different for each cases. The learning rates of tensor product parameters are one fifth of those of deep learning network parameters. For reproducing the original performance in \cite{hassani2021escaping}, the learning rate schedule of CCT / CVT remains unchanged, with the learning rate nonlinear declining. The learning rate schedules of LeNet-5, VGG, ResNet, and MobileNet with T-ADAF are listed in Table 1 and Table 2.}

\begin{table}[htbp]

\begin{tabular}{ccc}

\hline
  Epoch  & 1-50    & 51-100     \\ \hline
LeNet-5 & 0.1 & 0.02  \\ 
T-product & 0.02          & 0.004             \\ \hline
\end{tabular}
\caption{Learning Rate Schedule of LeNet-5 with T-ADAF }
\end{table}

\begin{table}[htbp]

\begin{tabular}{ccccc}

\hline
  Epoch  & 1-60    & 61-120  & 121 - 160 & 161-200  \\ \hline
VGG / ResNet / MobileNet & 0.1 & 0.02 & 0.004&0.0008 \\ 
T-product & 0.02          & 0.004        & 0.0008&  0.00016   \\ \hline
\end{tabular}
\caption{Learning Rate Schedule of VGG, ResNet and MobileNet with T-ADAF }
\end{table}

\begingroup
\setlength{\tabcolsep}{12pt}
\renewcommand{\arraystretch}{1.5}
\begin{table}[htbp]

\begin{tabular}{cccc}

\hline
    & $w_0$    & $w_1$    & $w_2$    \\ \hline
333 & $\frac{1}{3}$ & $\frac{1}{3}$ & $\frac{1}{3}$ \\ 
433 & 0.4           & 0.3           & 0.3           \\ 
525 & 0.5           & 0.25          & 0.25          \\ \hline
\end{tabular}
\caption{Suffixes of the model name and their meanings}
\end{table}

Here, $w_0$ is the wight of the original image. Essentially these parameters indicate how important the original image is relative to the other two images.

\subsection{Numerical Results of LeNet-5 with T-ADAF}
 In practical applications, especially with computational power and storage space limited, LeNet-5 is also an available model. This model, from the paper 
 \cite{lecun1998gradient}, is an old but very efficient convolutional neural network for handwritten character recognition. Meanwhile, algorithm scientists usually do not pursue an accuracy improvement of less than one percentage point, but are more concerned with reducing training costs. In order to better simulate the model evaluation metric in practical applications, two new definitions are introduced here.

\begin{definition}
\label{single_min}
For any single trained model $\mathcal F$, assume that its iteration sequence  is $\{t\in \mathbb N^+:1\leq t\leq T\}$, where $T$ is the maximum number of iterations. The corresponding training iteration parameter sequence is $\left\{\mathcal W_t\right\}_{1\leq t\leq T}$. Let $\left\{\lambda_t\right\}_{1\leq t\leq T}$ be the sequence of accuracy rates in the iterative process. Then the minimum available number of training iterations for the model $\mathcal F$ in this training iteration is defined as,
\begin{equation*}
t^{ava}=\argmin_{1\leq t\leq T}\left\{\lambda_t\geq\lfloor\lambda_{\max}\rfloor\right\},
\end{equation*}
where $\lfloor\cdot\rfloor$ means the round down operation and 
$$
\lambda_{\max}=\max_{1\leq t\leq T}\{\lambda_t\}.
$$
Meanwhile, the available accuracy rate is
$$
\lambda_{\min}^{ava}=\lfloor\lambda_{\max}\rfloor
$$
\end{definition}
\begin{definition}
\label{some_min}
A trained model group $\mathcal G$ contains one original model $\mathcal F_0$ and some other enhanced model $\mathcal F_i,i=1,2,\ldots,m$. Assume that their iteration sequences  are $\{t\in \mathbb N^+:1\leq t\leq T\}$, where $T$ is the maximum number of iterations.
Let $\left\{\lambda_t^{(i)}\right\}_{1\leq t\leq T}, i=0,1,\ldots,m,$ be the sequences of accuracy rates in these iterative processes.
 From Definition \ref{single_min}, the minimum available number of training iterations for the group $\mathcal G$ is defined as
\begin{equation*}
t^{ava}_i=\argmin_{1\leq t\leq T}\left\{\lambda_t^{(i)}\geq\lfloor\lambda_{\max}^{(0)}\rfloor\right\}.
\end{equation*}
for $i=1,2, \ldots ,m$.
\end{definition}

% In lots of practical problems, plenty of engineers have found that the model with hyper-parameters which have been adjusted before would not perform well in new cases. Therefore, we divide the whole numerical experiment into two parts, incompletely and completely pre-adjusting hyper-parameters, to avoid the influence of hyper-parameters for our framework. These two parts are depended on whether the hyper-parameters have been pre-adjusting for specific tasks. The hyper-parameters in VGG-Net and Resnet have not been pre-adjusted. However, in Vision Transformer, hyper-parameters have been adjusted yet. From this assumption, 

Here, we mainly test the performance of LeNet-5 on CIFAR-10 and CIFAR-100. The only thing to note is that LeNet-5 was originally designed to process greyscale images, while CIFAR10 and CIFAR100 are both RGB images. Therefore, we need to modify the parameters of the convolution kernel to make LeNet-5 meet our experimental requirements.

\begin{figure}[htbp]
	\centering
	\subfigure[CIFAR-10]{\includegraphics[width=2.8in]{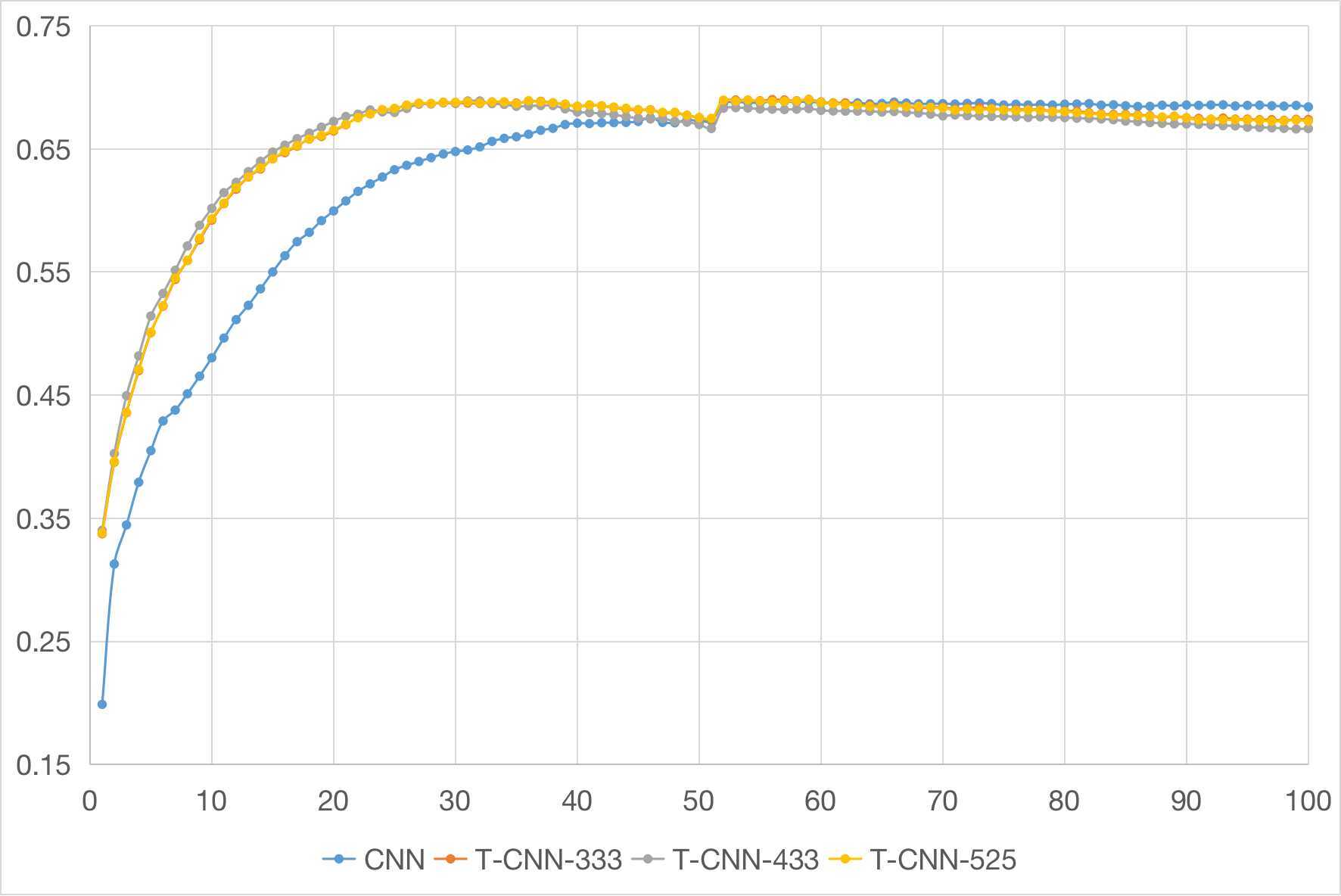}}
	\subfigure[CIFAR-100]{\includegraphics[width=2.8in]{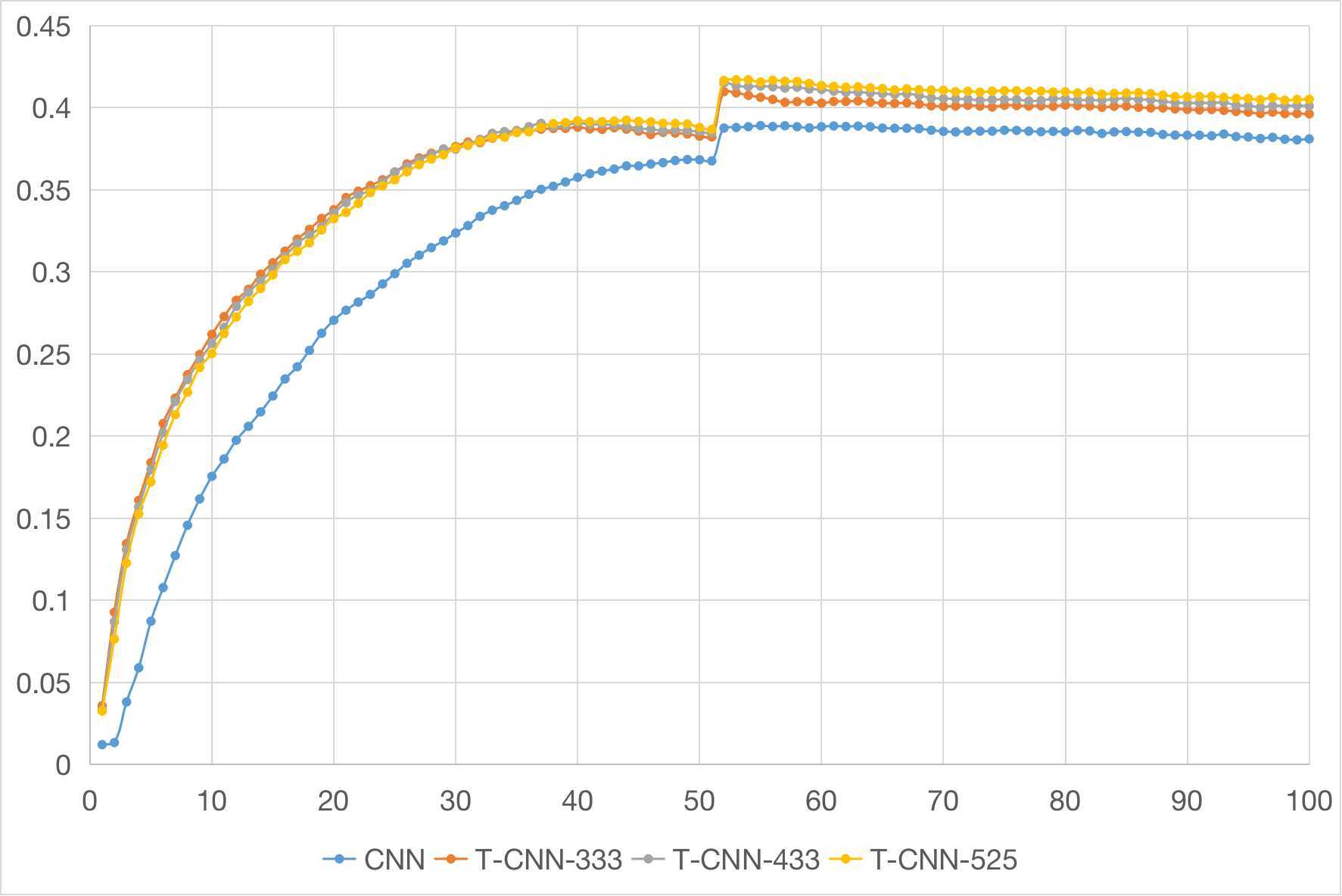}}
	\caption{LeNet-5 Model with T-product Enhancement Framework}
	\label{fig__5}
\end{figure}

{In the Figure \ref{fig__5}, we can see that for the CIFAR-100 dataset, the data augmentation framework can not only accelerate the iterative convergence speed of the deep learning model, but also improve the prediction accuracy of the original model. For the CIFAR-10 dataset, each category has 5000 samples to be learned, so the effect of final prediction accuracy enhancement is not obvious. But the data augmentation framework still reduces the number of iterations required. From Definition \ref{some_min}, we summarize the whole iteration processes as Table \ref{lambda_max}.}

\begin{table}[htbp]
\begin{tabular}{c|cc|ccc}
\hline
          &  $\lambda_{\max}^{(0)}$      &   $\lfloor\lambda_{\max}^{(0)}\rfloor$   &    $\lambda_{\max}^{(T333)}$    &    $\lambda_{\max}^{(T433)}$    &      $\lambda_{\max}^{(T525)}$ \\ \hline
CIFAR-10  & 0.6888 & 0.68 & 0.6876 & 0.6886 & \textbf{0.6898} \\
CIFAR-100 & 0.3859 & 0.38 & 0.4066 & 0.4118 & \textbf{0.4137} \\ \hline
\end{tabular}
\caption{Maximum Accuracy Rates of Numerical Results when Testing LeNet-5}
\label{lambda_ini}
\end{table}

\begin{table}[htbp]
\begin{tabular}{c|c|ccc}
\hline
          & Original & T333 & T433 & T525 \\ \hline
CIFAR-10  & 52       & \textbf{24}   & 24   & 25   \\
CIFAR-100 & 52       & 34   & \textbf{33}   & 35   \\ \hline
\end{tabular}
\caption{Minimum Epochs to Attend $\lfloor\lambda_{\max}^{(0)}\rfloor$ when Testing LeNet-5}
\label{lambda_max}
\end{table}

Among them, T433 represents the model that uses the T-product to enhance the framework and takes 0.4-0.3-0.3 as the weights. As it can be seen from Table \ref{lambda_ini} and Table \ref{lambda_max}, since the amount of data in the CIFAR-10 dataset is relatively sufficient, the advantages of the data enhancement framework are not obvious, and the effect of CIFAR-100 is more obvious. However, for both datasets, there is a clear advantage to reaching the minimum number of iterations available. This shows that in practical applications, a model using the data augmentation framework can be trained into use in fewer iteration steps.

\subsection{Numerical Results of Deep Neural Network with T-ADAF}
In this subsection, we test four models, VGG-16, ResNet-18, ResNet-34 and CCT-6, respectively. Due to the relatively high complexity and information extraction ability of these four models, the ten-category problem of the CIFAR-10 dataset is not challenging for them. The test accuracy rate has reached more than 95\%, even 98\%. Therefore, to better illustrate the performance of the adaptive data augmentation framework, we only show and analyze the performance of these four models.

%  All of these original models are set as the baselines in our experiments and the codes are available on GitHub \footnote[1]{The codes of VGG-16, ResNet-18 and ResNet-34 are available on https://github.com/weiaicunzai/pytorch-cifar100. The codes of CCT-6 are on https://github.com/SHI-Labs/Compact-Transformers}.

VGG-Net, first proposed in 2014, have proved that increasing the depth of the network can affect the final performance of the network. In VGG, multiple small convolution kernels are used instead of a large convolution kernel that had been commonly used before to reduce the calculation amount of convolution operations under the condition of ensuring the same receptive field. VGG-Net increases the depth of network and improves the effect of that. ResNet, which was first proposed in 2015 \cite{he2016deep}, used the residual structure in the neural network model for the first time, which solved the problem of gradient disappearance that often occurred when the number of layers in the previous model was increased. Since then, deep neural networks have become truly possible, and deep learning has gained more and more attention. Empirically, the depth of the network is critical to the performance of the model. When the number of network layers is increasing, the network can extract more complex feature patterns, so theoretically better results can be achieved. ResNet has gained good results in various competitions that year. Moreover, transformer \cite{devlin2018bert} was a common model in the field of natural language processing and was introduced in Computational Vision in 2018. Vision transformer (ViT) achieves further improvements in prediction performance by stacking attention modules \cite{vaswani2017attention}. CCT \cite{hassani2021escaping} reduces the number of parameters on the premise of ensuring the prediction accuracy, and provides convenience for the application and promotion of ViT.

\begin{figure}[htbp]
	\centering
	\subfigure[Epoch 10-200]{\includegraphics[width=2.8in]{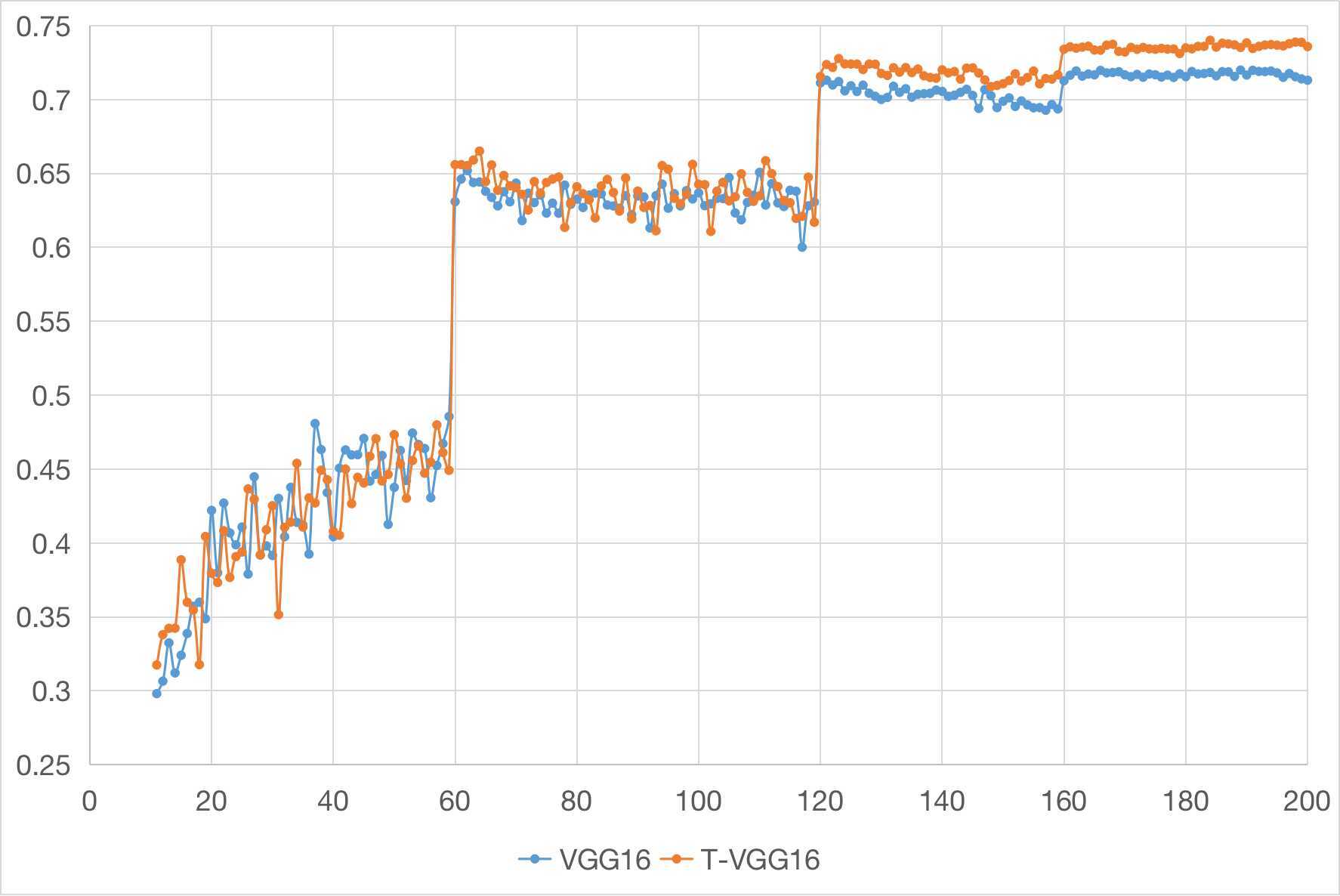}}
	\subfigure[Epoch 100-200]{\includegraphics[width=2.8in]{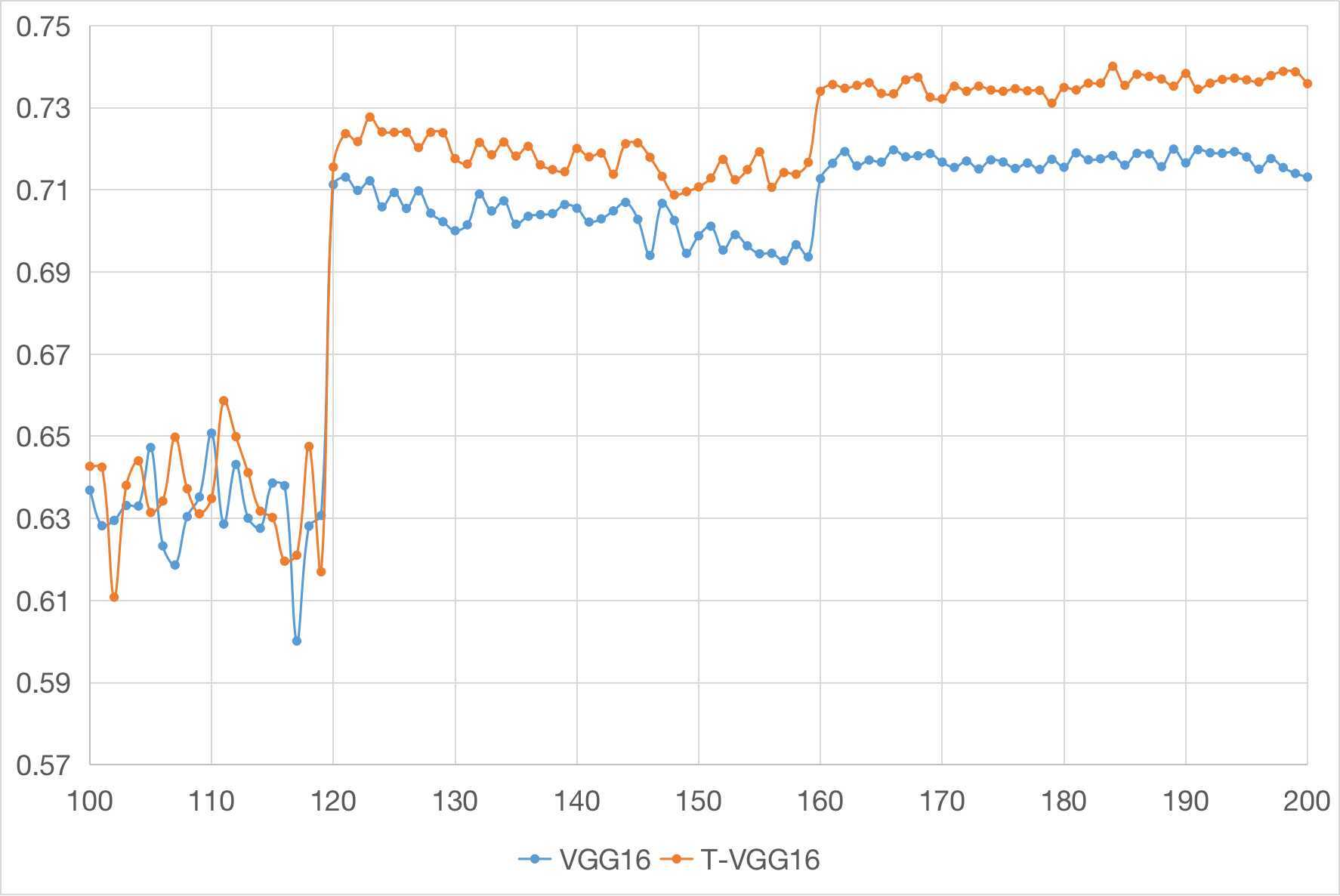}}
	\caption{VGG-16 Model with T-ADAF}
	%\label{fig1}
\end{figure}

\begin{figure}[htbp]
	\centering
	\subfigure[Epoch 10-200]{\includegraphics[width=2.8in]{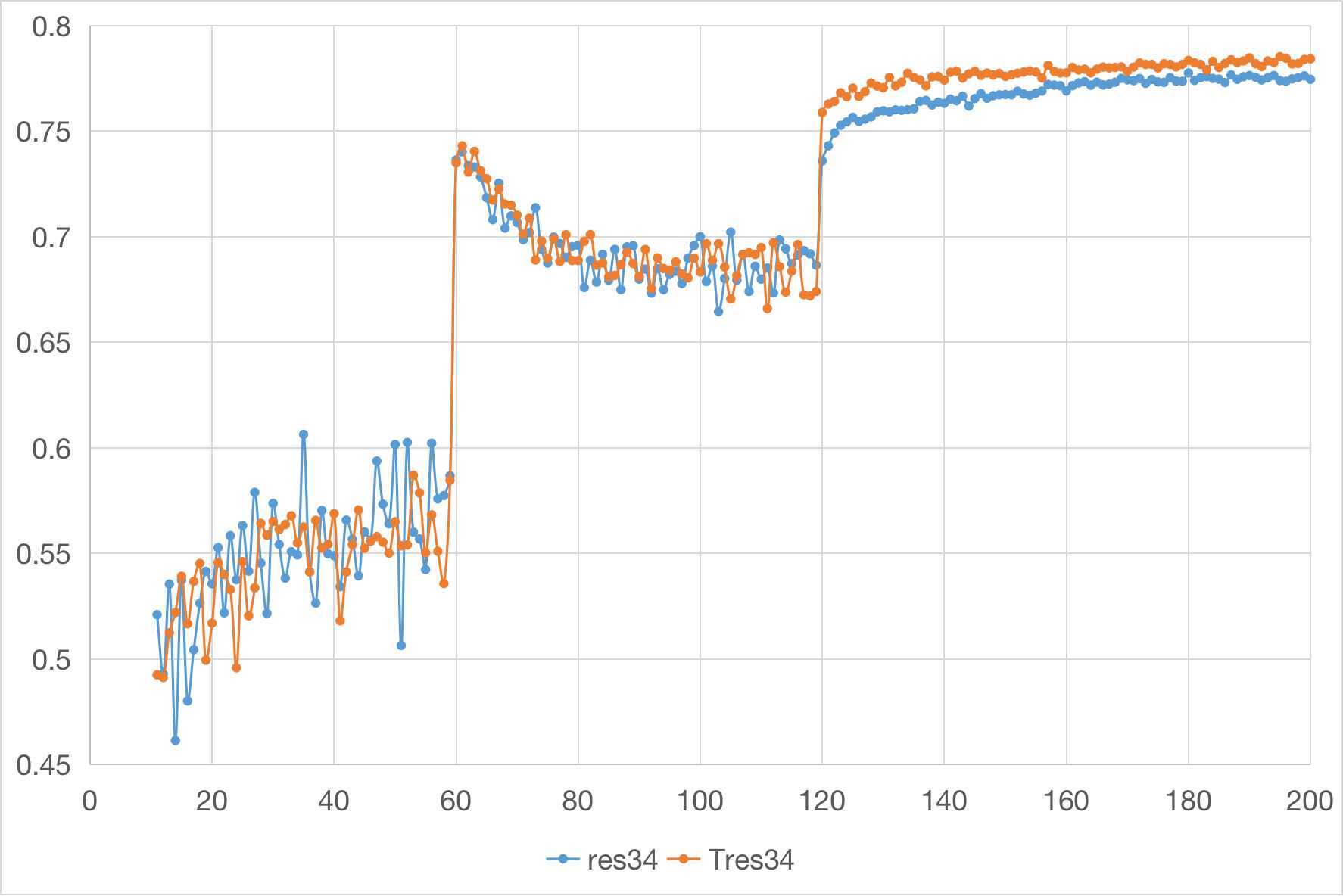}}
	\subfigure[Epoch 100-200]{\includegraphics[width=2.8in]{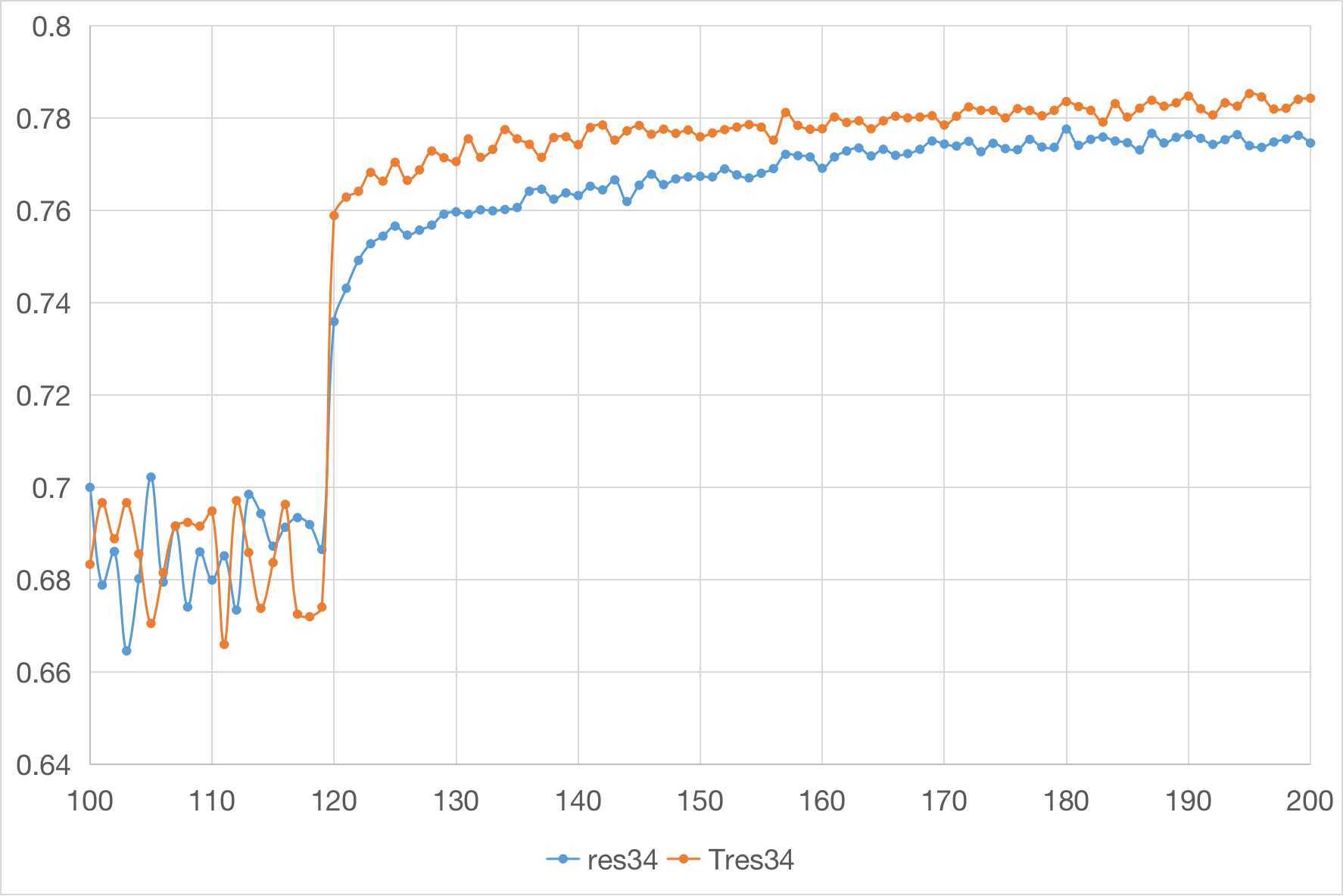}}
	\caption{ResNet-34 Model with T-ADAF}
	%\label{fig1}
\end{figure}

 \begin{figure}[htbp]
 	\centering
 	\subfigure[Epoch 10-200]{\includegraphics[width=2.8in]{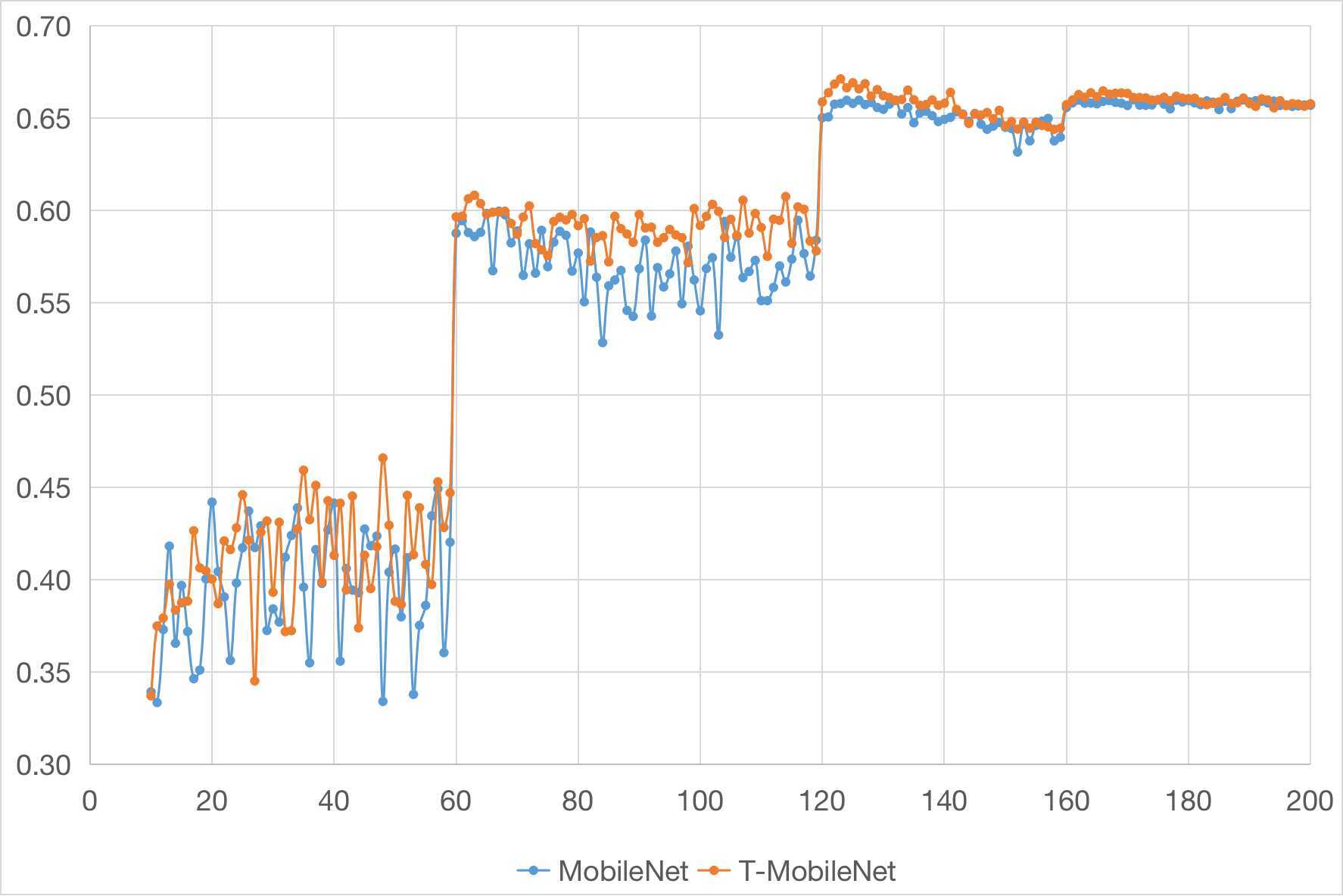}}
 	\subfigure[Epoch 100-200]{\includegraphics[width=2.8in]{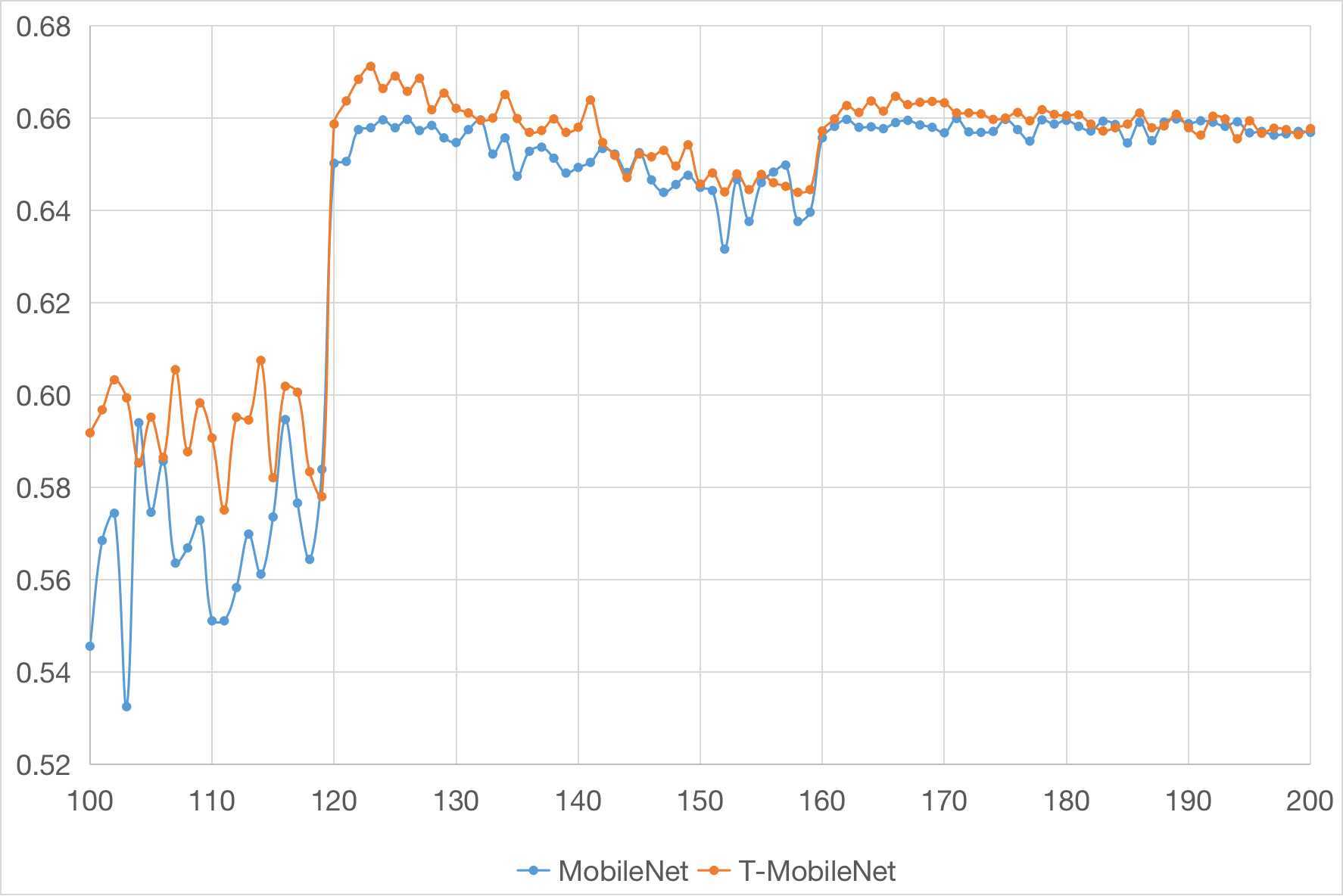}}
 	\caption{MobileNet Model with T-ADAF}
 	%\label{fig1}
 \end{figure}

\begin{figure}[htbp]
\centering
\includegraphics[width=5in]{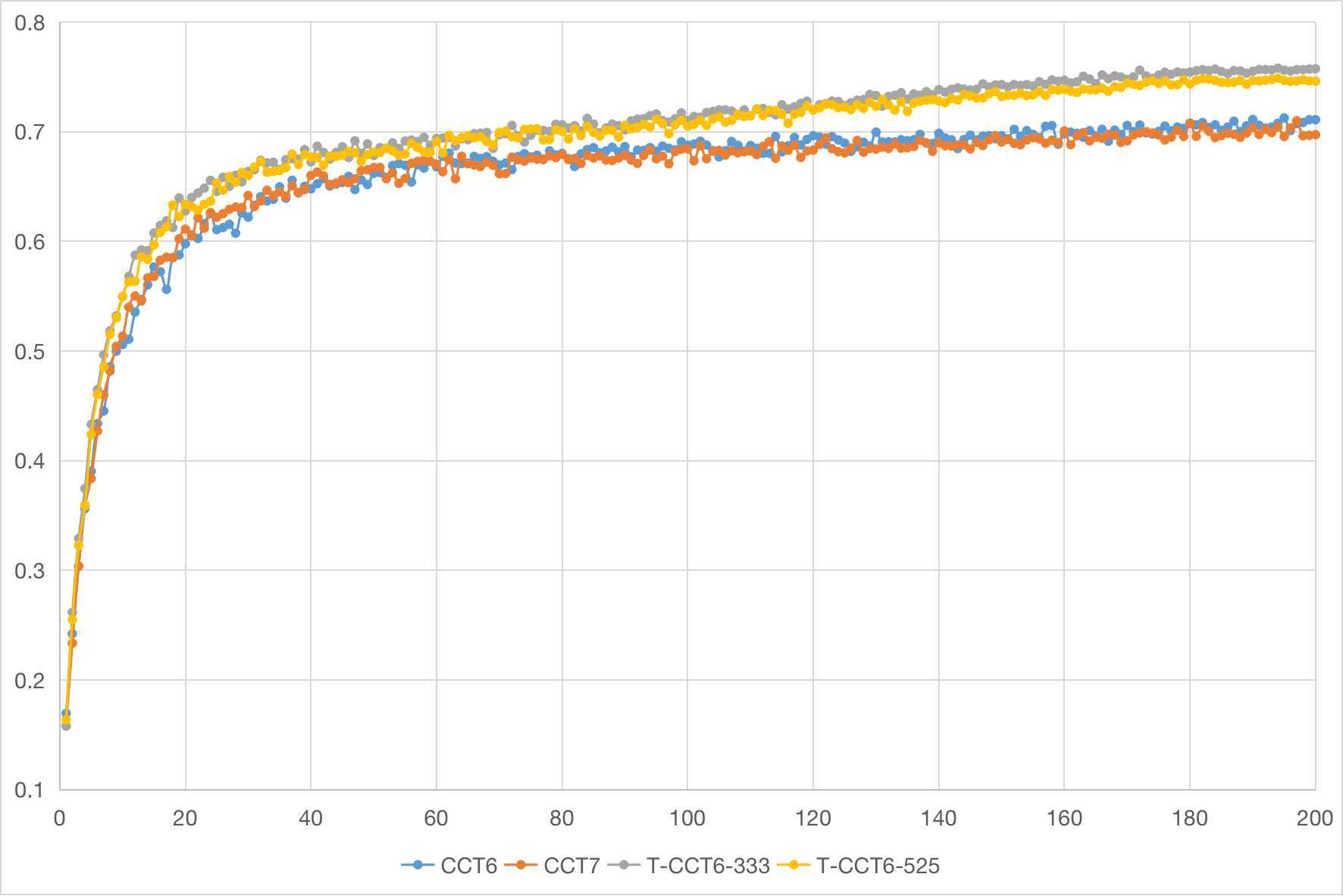}
\caption{CCT-6 Model with T-ADAF about Top-1 Accuracy on CIFAR-100}
\end{figure}
 
Figure 6-8 show how the prediction accuracy changes with the iterative epoch increasing. For convenience, we separately show the process of epoch 100-200 in the (b) picture. We choose VGG-16, ResNet-34 and CCT-6 as examples for showing iteration details.

\begin{figure}[htbp]
 \centering
 \includegraphics[width=5in]{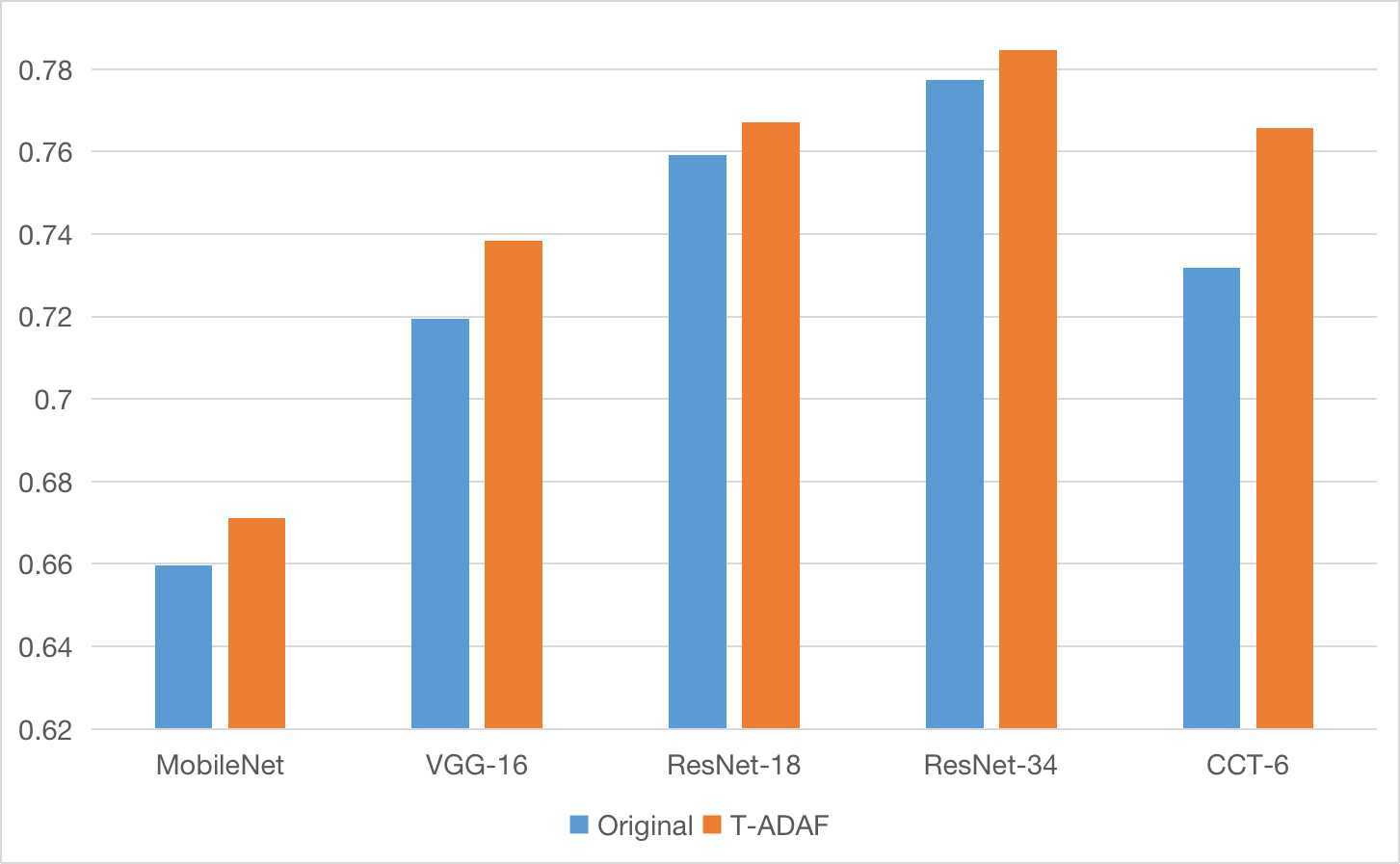}
 \caption{Comparisons among 5 Different Models and T-ADAF}
 \end{figure}
 
\setlength{\tabcolsep}{9pt}
\renewcommand{\arraystretch}{1}
\begin{table}[htbp]

\begin{tabular}{cccccc}
\hline
\multicolumn{2}{c}{\textbf{Model}}          & \textbf{Original} & \textbf{T-333}  & \textbf{T-433}  & \textbf{T-525}  \\ \hline
\multirow{2}{*}{\textbf{VGG-16}}    & Top-1 & 0.2805            & 0.2726          & 0.2674          & \textbf{0.2616} \\
                                    & Top-5 & 0.1061            & 0.0924          & 0.0921          & \textbf{0.0915} \\ \hline
\multirow{2}{*}{\textbf{ResNet-18}} & Top-1 & 0.2409            & 0.2356          & \textbf{0.2330} & 0.2361          \\
                                    & Top-5 & 0.0725            & 0.0648          & \textbf{0.0616} & 0.0652          \\ \hline
\multirow{2}{*}{\textbf{ResNet-34}} & Top-1 & 0.2226            & 0.2194          & \textbf{0.2153} & 0.2171          \\
                                    & Top-5 & 0.0601            & 0.0575          & \textbf{0.0558} & 0.0569          \\ \hline
\multirow{2}{*}{\textbf{CCT-6}}     & Top-1 & 0.2681            & \textbf{0.2342} & 0.2393          & 0.2405          \\
                                    & Top-5 & 0.0728            & \textbf{0.0621} & 0.0673          & 0.0714          \\ \hline
\end{tabular}

\caption{Errors of Numerical Results for Testing Deep Models on CIFAR-100}
\end{table}
From these results, we can see that the data augmentation framework can indeed improve the performance and fasten accelerate learning iteration speed of the model. The data augmentation frameworks under different weight settings are slightly different, but generally all perform better than the original models. 

\textbf{Remark:} 
\begin{itemize}
\item [(a)] In the adaptive data augmentation framework based on tensor T-product, the parameters in T-product in epoch 1-120 are learnable, while the T-product parameters are designed to remain unchanged in epochs 121-200. In addition, we have tested to find that fixing T-product parameter in epochs 121-200 does not affect the model performance and could avoid the risks of over-fitting.
\item [(b)]  Numerical experiments are based on Intel i5-10400F with NVIDIA 3080-10G. In the first 120 epochs, the running time of the enhanced model is about 4.5 times of that of the original model. Among them, the time consumption caused by data augmentation is three times, and the extra consumption caused by T-product operator is about 1.5 times of that of the original model. Meanwhile, in epochs 121-200, the enhanced model costs 2.2 times more time than the original model.
\end{itemize}
\subsection{Comparisons with Tensor Neural Network by Newman}

In 2018, Newman, Horesh, Avron and Kilmer \cite{newman2018stable} has proposed a tensor neural network based on T-product. What should be noted here is that the co-author, Kilmer, is an applied mathematician  who was the first scholar to come up with T-Product. Besides for the tensor neural network, Kilmer has received plenty of results in the filed, which combines tensor operators with the machine learning and some other applications \cite{zhang2014novel, soltani2016tensor, Newman2017Image}. 

In this subsection, we will introduce this network briefly and compare our adaptive data augmentation framework with Newman {\it et al.} \cite{newman2018stable}, in their paper, tensor forward propagation was defined as,
\begin{equation}
\mathcal A_{j+1} = \sigma(\mathcal W_j * \mathcal A_j+\mathcal B_j),
\end{equation}
for $j=0, 1,\ldots, N-1$, where $\sigma$ is an element-wise nonlinear activation function and $N$ is the numbers of layers in the network. We test MNIST and CIFAR-100 data sets on Tensor Neural Network and models with Adaptive Data Augmentation Framework based on T-product.
\setlength{\tabcolsep}{6pt}
\renewcommand{\arraystretch}{1}
\begin{table}[htbp]
\begin{tabular}{c|ccc}
\hline
         & Tensor Neural Network & LeNet-5 with T-ADAF & VGG-16 with T-ADAF \\ \hline
MNIST    & 97 - 98 \%              & 97.83 \%             & -                  \\
CIFAR-10 & 60 $\pm$ 0.5 \%         & 68.98 \%             & 91.25 \%            \\ \hline
\end{tabular}
\caption{Comparison of Tensor Neural Network and models with T-ADAF}
\end{table}

In the above table, we utilize T-ADAF as the short of Adaptive Data Augmentation Framework based on T-product. From this table, we can notice that T-ADAF improves the performance of the tensor neural network. 

\section{Summaries}
% In this section, we will analyze the Adaptive Data Augmentation Framework and its numerical experiments and gain some conclusions in summary.
% \subsection{Analysis}
In this paper, we propose an adaptive data augmentation framework by combining the tensor T-product and original neural network models. The RGB images are regarded as the third-order tensor data, and the T-product transformations of the two dimensions are performed in two dimensions of length and width. Such operations preserve the size of image and provide triple training samples for the learning of neural network model. 

As shown in the numerical experiments in the previous section, the T-product based data augmentation framework does have a good ability to augment the dataset and improve the model performance. At the same time, the proportion of parameters that are added is far smaller than that of improvements in model ability to acquire features. We use the \textbf{torchsummary} module in python to count model parameters. The parameters that we add for T-ADAF to be trained are two tensors $\mathcal W \in \mathbb{R}^{3\times 3\times 32}$. As a consequence, we only adopt 576 additional parameters to control the memory and computational power requirements.
\setlength{\tabcolsep}{6pt}
\renewcommand{\arraystretch}{1}

\begin{table}[htbp]
\begin{tabular}{c|ccc|ccc}
\hline
          & \makecell{Original \\Paramters} & \makecell{Additional \\Parameters} & \makecell{Additional \\ Proportion} & \makecell{Original \\Best-Acc} & \makecell{Enhanced \\Best-Acc} & Improvement \\ \hline
LeNet-5   & 145,578            & 576                   & \textbf{0.3957\%} & 38.59\%           & 41.37\%           & \textbf{2.78\%} \\
VGG-16    & 34,015,396         & 576                   & \textbf{0.0017\%} & 71.95\%           & 73.84\%           & \textbf{1.89\%} \\
ResNet-18 & 11,220,132         & 576                   & \textbf{0.0051\%} & 75.91\%           & 76.70\%           & \textbf{0.79\%} \\
ResNet-34 & 21,328,292         & 576                   & \textbf{0.0027\%} & 77.74\%           & 78.47\%           & \textbf{0.73\%} \\
CCT-6     & 3,717,733          & 576                   & \textbf{0.0156\%} & 73.19\%           & 76.58\%           & \textbf{3.39\%} \\ \hline
\end{tabular}
\caption{Comparison of Parameters and Best-Acc between Original models and Enhanced models with T-ADAF}
\label{result_sum}
\end{table}

From Table \ref{result_sum}, T-ADAF can improve the performance of original models with about 2\% on average. In 2021, Hassani \cite{hassani2021escaping} claims that their best Top-1 accuracy result of CCT model on CIFAR-100 is 75.59\%, while the best results of CCT with T-ADAF is 76.58\%. Moreover, the best results of ResNet-34 with T-ADAF is 78.47\%. It is worth noting that these results are derived directly from the training of CIFAR-100. Instead, better results can be obtained if transfer learning \cite{zhuang2020comprehensive} is performed using models that have been pre-trained on larger datasets.
% From the above table, indeed, the data augmentation framework based on T-product improves the ability of the model to extract feature information and accelerates the iterative convergence with a smaller number of added parameters. From our perspective, there are three main reasons:

 % \textbf{(1) Data Augmentation}.
 
% For a given model, the data-augmented framework triples one image tensor and gain the final result from three tensors together. At the same time, as an adaptive data augmentation framework, the mapping parameters of data are continuously learnable with iterations. This provides more intelligent data augmentation capabilities than traditional rigid body transform or similarity transform.

% \textbf{(2) Extracting more global features of image and their relationships}. 

% \textbf{(3) Image embedding to obtain more information from Rows and Columns}. 

% As shown in Figure 2, T-product can also be interpreted as performing FFT for tube data in two dimensions of length and width and then combining them together. This plays the role of embedding the image data in the same row and column. The vectors after embedding intersect each other, which is beneficial to extract deeper intersection information in the length and width dimensions.
%\subsection{Dataset Augmentation}
%% \subsection{Projections from different dimensions}
%\subsection{Image embedding to obtain more information from Rows and Columns}
%% 寻找更加global的图像内在联系
%\subsection{Extracting more global features of image and their relationships}
\subsection{Conclusions}

In one word, such a data augmentation framework has three main contributions as follows.

Firstly, the framework is an adaptive data augmentation framework which provides convenience for users. T-ADAF provides two more images for the original deep neural network model to gain the final prediction results together. Also, since the enhanced mapping parameters of the dataset are learnable, the framework will become more and more adaptable to the requirements of the problem as the training process going. It has obvious advantages over the more frequently used operations such as rotation and cropping.

Secondly, the framework greatly improves the model performance. The circulant convolution-like design in T-product improves the ability of the model to obtain global information. At the same time, the image embedding of length and width dimensions and the subsequent feature intersection also contribute to the prediction accuracy of the model. Numerical experiments show that the framework has great improvements and enhancements for both small and large models.

Finally, the framework combines tensor operators with neural networks, which provides deep learning models with the ability to process signals in both spatial and frequency domains. Meanwhile, T-Product module can obtain more global informations than CNN. The excellent mathematic properties of tensor T-product operators guarantee the performance of the model framework to a certain extent.

\subsection{Prospects of Future Research}
We believe that feature mining and machine learning based on high-order tensor data have broad research prospects. The relative position of data in three or higher dimensional space does contain plenty of information. In addition, it is also a meaningful topic to explain the effect of tensor operators on deep learning tasks from the perspective of numerical algebra.


\section*{Acknowledgment}

We  would like to thank the handling editor, three referees and
Dr. Jie Sun for their very useful comments on our paper.

\section*{Declaration of Interests}
The authors have not disclosed any competing interests.

\newpage



%\bibliographystyle{IEEEtran}

{\small
\bibliographystyle{unsrt}
\bibliography{ADAF-0404.bib}
}

%\begin{thebibliography}{1}
%\end{thebibliography}
%
%\appendix

\end{document}